\documentclass[journal,twoside]{IEEEtran}

\usepackage{moreverb,url}
\usepackage[colorlinks=black,bookmarksopen,bookmarksnumbered,citecolor=black,urlcolor=red]{hyperref}
\usepackage{graphicx}
\usepackage{psfrag}
\usepackage{bm}
\usepackage{subfigure}
\usepackage{multirow}
\usepackage{arydshln}
\usepackage{cite}
\usepackage{amsmath,amssymb}
\usepackage{xcolor}

\begin{document}

\title{Universal Flying Objects (UFOs): Modular Multirotor System for Flight of Rigid Objects}

\author{Bingguo~Mu,~\IEEEmembership{Student~Member,~IEEE,}
        and~Pakpong~Chirarattananon,~\IEEEmembership{Member,~IEEE}% 
%\thanks{Manuscript received April 19, 2005; revised August 26, 2015.}
\thanks{The authors are with the Department of Biomedical Engineering, City University of Hong Kong, Kowloon, Hong Kong SAR, China (e-mail: bingguomu2-c@my.cityu.edu.hk; pakpong.c@cityu.edu.hk).}% 
}

% The paper headers
\markboth{IEEE TRANSACTIONS ON XX,~Vol.~XX, No.~XX, Month~Year}{Mu \MakeLowercase{\textit{et al.}}: Universal Flying Objects (UFOs): Modular Multirotor System for Flight of Rigid Objects}%

\maketitle

\begin{abstract}
We introduce UFO, a modular aerial robotic platform for transforming a rigid object into a multirotor robot. To achieve this, we develop flight modules, in the form of a control module and propelling modules, that can be affixed to an object. The object, or payload, serves as the airframe of the vehicle. The modular design produces a highly versatile platform as it is reconfigurable by the addition or removal of flight modules, adjustment of the modules' arrangement, or change of payloads. To facilitate the flight control, we propose an IMU-based estimation strategy for rapid computation of the robot's configuration. When combined with the adaptive geometric controller for further refinement of uncertain parameters, stable flights are accomplished with minimal manual intervention or tuning required by a user. To this end, we demonstrate hovering and trajectory tracking flights through various robot's configurations with different dummy payloads, weighing $\approx200-800$ grams, using four to eight propelling modules. The results reveal that stable flights are attainable thanks to the proposed IMU-based estimation method. The flight performance is markedly improved over time through the adaptive scheme, with the position errors of a few centimeters after the parameter convergence. 

\end{abstract}


\section{Introduction}

\IEEEPARstart{M}{icro} Aerial Vehicles (MAVs) have gained remarkable popularity among scientists and engineers in recent years \cite{mahony2012multirotor,floreano2015science}. This has brought about significant progress in research and development in several related directions, including localization and mapping \cite{mur2017orb}, swarm robotics \cite{zhou2018agile,vasarhelyi2018optimized}, design and fabrication \cite{zhao2017deformable,pounds2018safety}, and dynamics and control \cite{ryll2015novel, mueller2016relaxed, antonelli2017adaptive}. These robots have demonstrated excellent versatility for a wide range of potential applications. When equipped with suitable tools, they become platforms for photography \cite{mcgarey2013autokite}, aerial manipulation \cite{abaunza2017dual}, grasping \cite{kessens2016versatile}, and delivery \cite{foehn2017fast, kim2018origami}. Together, multiple MAVs can cooperatively carry a suspended payload \cite{sreenath2013dynamics} or collaborate to open a door \cite{estrada2018forceful}.

In this work, we introduce UFO (Universal Flying Object)--a modular, reconfigurable, flying robotic system for rapid construction and incorporation with payloads or task-related components. In principles, these payloads can be simple rigid objects for delivery, or tools for manipulation. In this framework, we compartmentalize the robot into flight modules and payload. By treating the payload as the airframe of the robot, different robots can be constructed in various configurations from a combination of different payloads and flight modules depending on the intended application. With the developed standard flight modules (in the form of propelling and control modules) and the estimation and control algorithms, the framework facilitates the construction of different ready-to-fly platforms that are adeptly integrated with the payload while requiring minimal modeling and parameter tuning efforts from a user. In other words, we develop a robotic solution that provides a flight capability to the object of choice, akin to how the robotic skins in \cite{booth2018omniskins} turn inanimate objects into robots. 

To date, several robots have incorporated the modular architecture to benefit from the reconfigurability with the potential of improved robustness and lower costs \cite{seo2019modular, romanishin20153d}. In the domain of aerial robots, the Distributed Flight Array (DFA) has been proposed as a platform for research in distributed estimation and control algorithms \cite{oung2014distributed}. The ModQuad \cite{saldana2018modquad} is an individual flight-capable structure. In swarm, they cooperate and are capable of docking to create a flying structure midair. In another example, the transformable robot, DRAGON, exploits the multilink design to accomplish a multi-degree-of-freedom transformation for adapting to different environments \cite{zhao2018design}. It can be seen that, the modular design concept present in these robots, compared to our proposed platform, serves radically different purposes. In our design, the modularity can be categorized as having a ``slot" architecture according to \cite{seo2019modular} as not all modules are interchangeable, whereas the mentioned examples, to large extent, feature a ``sectional" architecture such that there exists no base component. This reflects the contribution of our work, that is to create a flying platform that integrates with payloads, allowing a single platform to be easily used across a wide range of applications.

\section{System Overview}\label{sec.systemdescription}
\subsection{Modular design}
\begin{figure}[t!]
\centering
\psfrag{system}[c][c][0.9]{multirotor system}
\psfrag{cmo}[c][c][0.9]{control module}
\psfrag{pmo}[c][c][0.9]{propelling module}
\psfrag{payload}[c][c][0.9]{payload}
\psfrag{N}[c][c][0.9]{N}
\includegraphics[width=73mm]{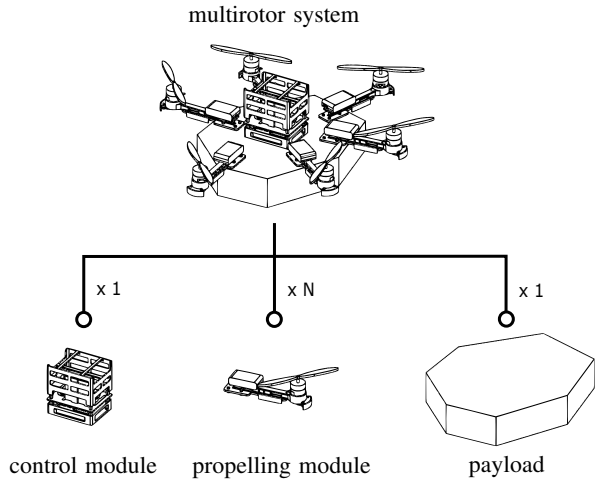}
\caption{The proposed modular system is composed of one control module, multiple propelling modules and an object (payload/airframe).}\label{fig.UFOs_concept}
\end{figure}

We propose an aerial robotic platform with a modular design. The UFO consists of (i) multiple ($N$) propelling modules (the dynamic requirements impose the constraint: $N\geq4$, while the maximum $N$ is limited by practicality.), (ii) a control module, and (iii) a payload.  To create a flight capable device, the propelling modules and control module are affixed to the rigid payload, which serves as a structural component or an airframe, forming a vehicle shown in Fig. \ref{fig.UFOs_concept}. 

The propelling module, consisting primarily of a motor, an electronic speed controller, a battery and a propeller, are responsible for generating thrust and torque as commanded by the control module. The control module houses an onboard flight controller, a battery and necessary sensors required for flight stabilization and control, similar to conventional multirotor robots. Both propelling modules and control module, also referred to as flight modules, feature a mechanism for attachment with the payload. A different number of propelling modules can be used to construct a vehicle based on the weight and size of the payload. That is, a lightweight payload requires the minimum of four propelling modules, whereas a heavier payload can be transported with six, eight, or more propelling modules. This modular concept is beneficial as it allows different robots to be made of the same set of hardware (propelling and control modules), using the same control strategy. 

Since our modular design integrates the payload as the structural component of the robot, this inevitably leads to certain restrictions on the rigidity of the payload. The geometry and surface material of the payload must also facilitate physical attachments of the modules. The arrangement of these modules must also be compatible with the dynamic constraints, such that the singularity condition is avoided. In other words, the robots must be able to generate torque about the three body axes independently. In addition, to simplify the control strategy, all propelling thrusts must be aligned (similar to conventional multirotor systems). This requires the payload to have a flat surface for attachment of flight modules.

Owing to the reconfigurable nature of the system, we must overcome the subsequent difficulties related to flight control and stabilization. Traditional flight controllers rely on prior knowledge of the mass distribution of the robot and locations of the propellers for calculation of body torque. The controller is then designed and controller gains are tuned for each respective robot. In our case, since the modules are to be paired with a payload with unknown physical properties, the flight controller must be able to handle the lack of prior information about the system.

The issue of unknown system parameters is tackled in two steps. The first step is to obtain initial estimates of the parameters. Due to several limitations, these estimates are inaccurate and bound to affect the flight performance adversely. In the second step, we devise an adaptive flight controller that is capable of dealing with a system with a large number of uncertain parameters to further refine the initial estimates and improve the flight performance.

To compute initial system parameters, which include the mass, moment of inertia, and locations of the propellers with respect to the center of mass (CM) of the robot, we incorporate an inertial measurement unit (IMU) into each flight module. Under the assumption that all propelling thrusts are approximately aligned, the measurements from the IMUs are used to deduce the relative distances between all the flight modules. Critical parameters required for flight are then estimated based on this knowledge. 

Since these estimates are inevitably inaccurate, they are regarded as initial estimates for flight control purposes. While they may be sufficiently accurate for the robots to stabilize and stay airborne, the flight performance must be further improved for practical uses. To this end, we develop an adaptive geometric flight controller that is capable of refining a large number of uncertain robot parameters. The stability and convergence of the proposed flight controller is provided via a Lyapunov analysis. 

Unlike a conventional approach where a gripper is often used for aerial transport \cite{kim2018origami,estrada2018forceful}, our proposed modular design provides an alternative solution. While a gripper-based method usually does not require human intervention in picking up objects, the proposed method leverages the user's involvement for attaching modules to the payload and performing manual calibration. With the proposed modular design, our proposed multirotor system has a reconfigurable structure for uses with different payloads as the airframe.  The reconfigurable property comes with several associated challenges related to flight control and stability. This is addressed by the IMU-based parameter estimation strategy and a flight controller that can adaptively and comprehensively improve the flight performance over time.

\subsection{Related work and technical contribution}

In addition to the novelty in the proposed modular design and its potential applications, the contribution of this paper extends to (i) the parameter estimation strategy; and (ii) the flight control method.

As mentioned, we employ multiple IMUs on the robot for estimation of the relative distance and orientation between flight modules. To date, there have been few examples  where multiple IMUs are deployed on a flying robot. In \cite{guerrier2012fault, avram2015imu}, several IMUs were used for fault detection and estimation of sensor bias. In the previously developed modular flying robots \cite{oung2014distributed, saldana2018modquad}, the configuration of the robot is determined in advance, the IMU on each module is used for the distributed flight control tasks. In our design, IMUs on all flight modules are used for the estimation of the robot's configuration, but only the IMU from the control module is used for flight control. The use of multiple IMUs here is more akin to
\cite{rehder2016extending}, where the authors developed a method for spatially calibrating multiple IMUs and a camera.

In terms of control, several flight controllers suitable for multirotor systems have been proposed. A number of seminal works rely on the Euler angles or quaternions to represent the attitude dynamics in an attempt to address the inherent nonlinearity of the systems \cite{mellinger2012trajectory,faessler2015automatic}. The nonlinear approach offers benefits over the linearized methods as they are capable of tracking more aggressive trajectories. Inevitably, these come with the associated singularities or ambiguity. Therefore, more recent developments have explored the global expression of the special Euclidean group, SE(3) \cite{goodarzi2013geometric, faessler2017thrust}. To deal with parameter uncertainties adaptive laws have been incorporated \cite{goodarzi2015geometric, chirarattananon2016perching, mu2017adaptive}.

The adaptive geometric flight controller in this work is a notable extension from \cite{chirarattananon2016perching,mu2017adaptive}. The proposed controller, in a similar fashion to \cite{goodarzi2013geometric, faessler2017thrust}, directly controls the robot on SE(3). In the meantime, the adaptive law is capable of updating a large set of uncertain parameters, caused by the unknown robot's configuration. The stability of the highly nonlinear dynamics is given by the Lyapunov analysis with a few simplifying assumptions. The nonlinear approach is also suitable for more aggressive maneuvers if required.

The derivation of the nonlinear controller here leverages the fact that the attitude dynamics of the robot are related to the higher-order components of the translational dynamics. As a result, we simultaneously consider both attitude and translational dynamics for the position control. Unlike the implementation in \cite{goodarzi2015geometric}, this eliminates the need to determine the attitude and angular velocity setpoints as an intermediate step, permitting us to directly use the position and yaw angle--the flat outputs of the system--as the setpoints at the cost of a simplifying assumption related to the altitude and thrust of the robot.

\section{Dynamics Model} \label{sec.systemmodel}

The flight dynamics of the proposed modular system are fundamentally similar to those of conventional multirotor vehicles, with the exception that many important parameters are not \textit{a priori} known and have to be estimated.

\subsection{Flight dynamics}
\begin{figure}[ht]
\centering
\psfrag{xw}[c][c][0.9]{$X_\text{W}$}
\psfrag{Yw}[c][c][0.9]{$Y_\text{W}$}
\psfrag{zw}[c][c][0.9]{$Z_\text{W}$}
\psfrag{xb}[c][c][0.9]{$X_\text{B}$}
\psfrag{Yb}[c][c][0.9]{$Y_\text{B}$}
\psfrag{zb}[c][c][0.9]{$Z_\text{B}$}
\psfrag{xp}[c][c][0.9]{$X_{P_i}$}
\psfrag{Yp}[c][c][0.9]{$Y_{P_i}$}
\psfrag{zp}[c][c][0.9]{$Z_{P_i}$}
\psfrag{system}[c][c][0.9]{body frame}
\psfrag{world}[c][c][0.9]{inertial frame}
\psfrag{module}[c][c][0.9]{module frame}
\psfrag{Ti}[c][c][0.9]{$T_i$}
\psfrag{P}[c][c][0.9]{$\bm{P}$}
\psfrag{T}[c][c][0.9]{$T$}
\psfrag{mg}[c][c][0.9]{$mg$}
\psfrag{r_i}[c][c][0.9]{$r^\text{cm}_i$}
\includegraphics[width=80mm]{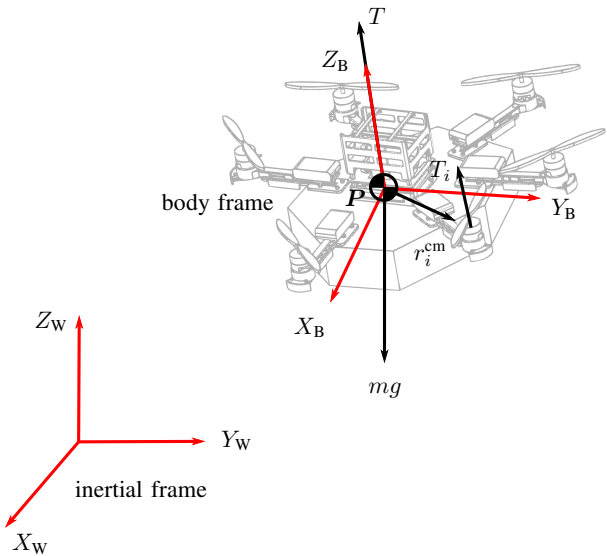}
\caption{A flying vehicle with the defined reference frames.}\label{fig.fundamentals}
\end{figure}
The flight dynamics are described with respect to two reference frames as shown in Fig. \ref{fig.fundamentals}. The inertial frame \{$X_\text{W}$, $Y_\text{W}$, $Z_\text{W}$\} is fixed while the body frame \{$X_\text{B}$, $Y_\text{B}$, $Z_\text{B}$\} is assumed located at the center of mass (CM) of the robot. A rotation matrix $\bm{R}\in \text{SO(3)}$ maps the body frame to the inertial frame. This rotation matrix can also be expressed in the form of column vectors as $\bm{R}=[\hat{\bm{i}}, \hat{\bm{j}}, \hat{\bm{k}}]$, where $\hat{\bm{i}}$, $\hat{\bm{j}}$, $\hat{\bm{k}}$ represent the corresponding vectors of three body axes in the inertial frame. Let $\bm{\omega}=[\bm{\omega}_x, \bm{\omega}_y, \bm{\omega}_z]^\text{T}$ denote the angular velocity, then the time derivative of the rotation matrix is given as
\begin{equation}
\dot{\bm{R}}=\bm{R}\bm{\omega}_\times,\label{eqn.rotationmatrix}
\end{equation}
where $\bm{\omega}_\times$ is the skew-symmetric matrix representation of $\bm{\omega}$. The rotational dynamics of the robot depends on the total torque $\bm{\tau}$ acting on the system:
\begin{equation}
\bm{\tau}=\bm{I}\dot{\bm{\omega}}+\bm{\omega}\times\bm{I}\bm{\omega},
\end{equation}
where $\bm{I}$ is the inertia tensor of the system.

Let $\bm{P}=[x,y,z]^\text{T}$ denote the CM position of the robot in inertial frame and $m$ be the total mass, the translational dynamics of the robot are
\begin{equation}
\ddot{\bm{P}}=\frac{T}{m}\hat{\bm{k}}+\bm{g},
\label{eqn.positions}
\end{equation}
where $\bm{g}=[0,0,-g]^\text{T}$ is the gravity vector, and $T$ is the total force produced by the propelling modules, assumed to be in the direction of $\hat{\bm{k}}$. We simplify our analysis to a near-hovering condition ($\bm{\omega}\rightarrow \bm{0}$), the term $\bm{\omega}\times\bm{I}\bm{\omega}$ can be neglected. The rotational and translational dynamics are consolidated as
\begin{equation}
\left[
  \begin{array}{c}
    m(\ddot{z}+g)\\
    \bm{I}\dot{\bm{\omega}}
  \end{array}
\right]
\approx
\left[
  \begin{array}{c}
    T\\
    \bm{\tau}
  \end{array}
\right].
\label{eqn.thrusttorquesimplified}
\end{equation}
The quantity on the right hand side of equation \eqref{eqn.thrusttorquesimplified} are the total thrust and moment generated by the propellers. The total thrust is a summation of the force produced by individual propeller, $T=\sum_{i=1}^{N}T_i$. The resultant roll and pitch torques depend on the location of each propeller with respect to the CM, $\bm{r}^{\text{cm}}_i=[\bm{r}^{\text{cm}}_{ix},\bm{r}^{\text{cm}}_{iy},\bm{r}^{\text{cm}}_{iz}]^\text{T}$ as illustrated in Fig. \ref{fig.fundamentals}. The aerodynamic drag from the propellers contribute to the yaw torque. The thrust and torque of each propeller are assumed to be quadratic functions of the spinning rate of the propeller \cite{oung2014distributed}. Defining $c_i$ as an aerodynamic constant representing the ratio of drag torque to thrust generated from the $i^\text{th}$ propeller ($c_i$ is either positive or negative, depending on the spinning direction of the propeller), we obtain
\begin{equation}
\left[
  \begin{array}{c}
    T\\
    \bm{\tau}
  \end{array}
\right]
=
\underbrace{
\left[
  \begin{array}{ccccc}
    1 & 1 & \cdots & 1 \\
    \bm{r}^{\text{cm}}_{1y} & \bm{r}^{\text{cm}}_{2y} & \cdots & \bm{r}^{\text{cm}}_{Ny} \\
    -\bm{r}^{\text{cm}}_{1x} & -\bm{r}^{\text{cm}}_{2x} & \cdots & -\bm{r}^{\text{cm}}_{Nx} \\
    c_1 & c_2 &\cdots & c_N \\
  \end{array}
\right]
}_{\bm{A}_u}
\underbrace{
\left[
  \begin{array}{c}
    T_1\\
    T_2\\
    \vdots \\
    T_{N} \\
  \end{array}
\right]
}_{\bm{u}}.
\label{eqn.stateAprime}
\end{equation}
In equation \eqref{eqn.stateAprime}, we treat the propelling thrusts as the system's input $\bm{u}$. The matrix $\bm{A}_u$ relates the input $\bm{u}$ to the total thrust and moment of the robot. In this work, since we assume that the modular system is reconfigurable by a user, all elements in the second and third rows of $\bm{A}_u$ are unknown. In contrast, the values of $c_i$ are predetermined.

Combining equation \eqref{eqn.stateAprime} with the system dynamics from equation \eqref{eqn.thrusttorquesimplified} yields
\begin{equation}
\begin{split}
\left[
  \begin{array}{c}
    \ddot{z}+g\\
    \dot{\bm{\omega}} \\
  \end{array}
\right]
&=
\left[
  \begin{array}{cccc}
    m & \bm{0}_{1\times3} \\
    \bm{0}_{3\times1} & \bm{I} \\
  \end{array}
\right]^{-1}\bm{A}_u\bm{u}=\bm{A}\bm{u},
\end{split}\label{eqn.dynamicmodel2}
\end{equation}
where $\bm{A}\in\mathbb{R}^{4\times N}$ is the configuration matrix. The matrix  $\bm{A}$ is also unknown as $\bm{A}_u$, $m$, and $\bm{I}$ are not pre-determined because they depend on the payload and the locations of the flight modules. 

\section{IMU-based Parameter Estimation Methods}\label{sec.imu-based_estimation}

To achieve a stable flight, we first estimate the configuration matrix $\bm{A}$ in equation \eqref{eqn.dynamicmodel2}. This is accomplished with the help of multiple IMUs located on all flight modules. The estimate of $\bm{A}$ is computed based on the results from three steps: estimations of total mass $m$, moment arms $\bm{r}^{\text{cm}}_{i}$'s, and inertia tensor $\bm{I}$.

\subsection{Estimation of total mass}

Let $m_{i}$ and $m_\text{p}$ denote the mass of the $i^\text{th}$ module and the mass of the payload, respectively. The total mass of the robot, $m$, is 
\begin{equation}
m=\sum^{N}_{i=0}m_{i}+m_\text{p},\label{eqn.mass_estimation}
\end{equation}
where $m_{0}$ is the mass of the control module, and $m_{i}, i=1,\ldots,N$, is the mass of the propelling module. The values of $m_i$ are known, but the payload mass $m_\text{p}$ is unknown. Therefore, $m_\text{p}$ is assumed to scale with $N$ such that $m_\text{p}=m_{n} N$, where $m_{n}$ is an assumed payload mass normalized by the number of propelling modules. The value of $m_{n}$ depends on the payload capability of each propelling module. This is a reasonable assumption since the number of the propelling modules required depends on the mass of the payload and a user can approximate the number of modules needed when constructing the robot. If the payload is too heavy for a liftoff, more propelling modules can be added.

\subsection{Estimation of moment arms}

According to equation (\ref{eqn.stateAprime}), roll and pitch torques are dictated by the length of the moment arms, $\bm{r}^\text{cm}_{i}$, shown in Fig. \ref{fig.forinertiatensor}. Here, we propose a strategy to use IMU measurements from all ($N+1$) flight modules to estimate the locations $\bm{r}^\text{cm}_{i}$ of the rotors.

With no prior knowledge of the CM location of the payload, the moment arm $\bm{r}^\text{cm}_{i}$ of each rotor can not be directly determined. Since the propelling modules must be placed on the edge of the payload, it is fair to assume that, on the $X_\text{B}$-$Y_\text{B}$ plane, the CM is situated close to the geometric center (GC) of all $N+1$ flight modules, or $\bm{r}^\text{cm}_{i}\approx\bm{r}^\text{gc}_{i}$. Consequently, we seek to estimate $\bm{r}^\text{gc}_{i}$ instead.

By design, in our current hardware, the IMUs on propelling modules are installed directly below the rotors. In the $X_\text{B}$-$Y_\text{B}$ plane, the location of the IMU conveniently represents the location of the rotor. The orientation of these IMUs are not necessarily aligned. Without loss of generality, we use the orientation of the IMU belonging to the control module to represent the orientation of the body frame.

Let $\bm{R}_{i}$ be a rotation matrix that maps the frame of the $i^\text{th}$ IMU to the body frame (or the $0^\text{th}$ IMU) and $\bm{\omega}_i$ be the gyroscopic reading from the $i^\text{th}$ IMU. Then, it is expected that
\begin{equation}
\bm{R}_{i}\bm{\omega}_{i}=\bm{\omega}_{0}=\bm{\omega}.\label{eqn.modulerationmatrix}  
\end{equation}
With multiple measurements over time, we transform equation \eqref{eqn.modulerationmatrix} by stacking the elements of $\bm{R}_{i}$ into a column vector to solve for $\bm{R}_{i}$ via least-squares method. The solution is numerically scaled to impose the special orthogonality condition.

To obtain $\bm{r}^\text{gc}_{i}$, we consider the relative position between the GC $(\bm{P}_\text{gc})$ and the $i^\text{th}$ IMU $(\bm{P}_i)$ in the inertial frame: $\bm{P}_i=\bm{P}_\text{gc}+\bm{R}\bm{r}^\text{gc}_{i}$. It follows that $\ddot{\bm{P}}_i=\ddot{\bm{P}}_\text{gc}+\ddot{\bm{R}}\bm{r}^\text{gc}_{i}$. The $i^\text{th}$ IMU provides the reading of the gravity-subtracted acceleration: $\bm{a}_i=\bm{R}^\text{T}_{i}\bm{R}^\text{T}(\ddot{\bm{P}}_i-\bm{g})$. Together, this yields
\begin{equation}
\bm{R}\bm{R}_{i}\bm{a}_{i}+\bm{g}=\ddot{\bm{P}}_\text{gc}+\ddot{\bm{R}}\bm{r}^\text{gc}_{i}.\label{eqn.imu0}
\end{equation}
Next, we sum equation \eqref{eqn.imu0} over all $N+1$ IMUs. With a simple re-arrangement, this becomes
\begin{equation}
\sum^{N}_{i=0}(\bm{R}_{i}\bm{a}_{i})+(N+1)\bm{R}^\text{T}(\bm{g}-\ddot{\bm{P}}_\text{gc}) =\bm{R}^\text{T}\ddot{\bm{R}}\sum^{N}_{i=0}\bm{r}^\text{gc}_{i},\label{eqn.imu1}
\end{equation}
where, by definition, $\sum^{N}_{i=0}\bm{r}^\text{gc}_{i}=0$. We substitute equation \eqref{eqn.imu0} into equation \eqref{eqn.imu1} to get rid of the $\bm{g}-\ddot{\bm{P}}_\text{gc}$ term:
\begin{equation}
\bm{R}^\text{T}\ddot{\bm{R}}\bm{r}^\text{gc}_{i}=\bm{R}_{i}\bm{a}_{i}-\frac{1}{N+1}\sum^{N}_{i=0}(\bm{R}_{i}\bm{a}_{i}).\label{eqn.imualgorithmp}
\end{equation}
According to equation \eqref{eqn.rotationmatrix}, the term $\bm{R}^\text{T}\ddot{\bm{R}}$ from equation \eqref{eqn.imualgorithmp} can be computed from $\bm{\omega}$ as
\begin{equation}
\bm{R}^\text{T}\ddot{\bm{R}}=\bm{\omega}_\times \bm{\omega}_\times+\dot{\bm{\omega}}_\times. \label{eqn.imu2}
\end{equation}
To attenuate the measurement noise, instead of relying on a single IMU for $\bm{\omega}$, we opt to use the averaged measurements from all IMUs, $\bar{\bm{\omega}}$, given as
\begin{equation}
\bar{\bm{\omega}}=\frac{1}{N+1}\sum^{N}_{i=0}\bm{R}_{i}\bm{\omega}_{i}, \label{eqn.averagedangularvelocity}
\end{equation}
Finally, equation \eqref{eqn.imualgorithmp} becomes
\begin{equation}
(\bar{\bm{\omega}}_\times \bar{\bm{\omega}}_\times+\dot{\bar{\bm{\omega}}}_\times)\bm{r}^\text{gc}_{i}=\bm{R}_{i}\bm{a}_{i}-\frac{1}{N+1}\sum^{N}_{i=0}(\bm{R}_{i}\bm{a}_{i}).\label{eqn.imualgorithm}
\end{equation}
Then, it is straightforward to solve for $\bm{r}^\text{gc}_{i}$ via the linear least-squares method, using the measurements from all the IMUs. As stated, we assume $\bm{r}^\text{cm}_{i}\approx \bm{r}^\text{gc}_{i}$ for the rest of the paper.

\subsection{Estimation of inertia tensor} \label{sec.est_inertia}

To estimate the inertia tensor of the robot ($\bm{I}$), we let $\bm{I}_{i}$  and $\bm{I}_\text{p}$ denote the inertia tensor of the $i^\text{th}$ module and the inertia of the payload, both defined with respect to the CM of the robot. The total inertia is the sum of all components:
\begin{equation}
\bm{I}=\sum^{N}_{i=0}\bm{I}_{i}+\bm{I}_\text{p},\label{eqn.inertia_estimation0}
\end{equation}
\begin{figure}[t]
\centering
\psfrag{ri}[c][c][0.9]{$\bm{r}^{\text{cm}}_{i}$}
\includegraphics[width=60mm]{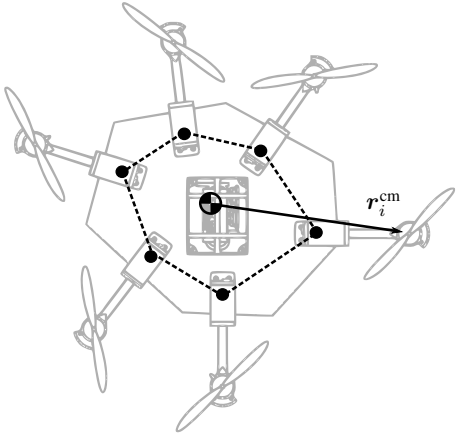}
\caption{The module system with the flight modules and the payload. The mounting points are indicated by the black dots and the dashed lines represent the estimated shape of the payload.}\label{fig.forinertiatensor}
\end{figure}

Since the control module and propelling modules are custom-made, the estimates of their inertia tensors with respect to their CM of the modules are available from the computer aided design software (CAD). With the estimates of the moment arms ($\bm{r}^\text{cm}_i$) and the orientation with respect to the CM of the entire robot ($\bm{R}_i$) from above, we immediately obtain the location of the CM of the module in the body frame. Then, the parallel axis theorem is used to compute $\bm{I}_{i}$, the inertia tensor of the $i^\text{th}$ module in the body frame.

Without direct measurements, a few simplifying assumptions are required to estimate the inertia tensor of the payload, $\bm{I}_\text{p}$. Here, we treat the payload as an $N$-sided infinitesimally thin, flat polygon with uniform density. These assumptions conveniently allow us to compute $\bm{I}_\text{p}$ based on only previously estimated parameters, such as $m_\text{p}$, $\bm{r}_i^\text{cm}$, and $\bm{R}_i$. 
To evaluate the inertia tensor of an $N$-sided polygon with respect to the CM of the robot, we assume the vertices are located at the mounting point of the propeller modules shown in Fig. \ref{fig.forinertiatensor}. These mounting points, in the body frame, can be determined from $\bm{r}_i^\text{cm}$, $\bm{R}_i$, and the physical design of the module.

\subsection{Estimation of the configuration matrix}

With the estimates of the total mass, the moment arms, and the moment of inertia of the robot, the matrix $\bm{A}_u$ and the configuration matrix $\bm{A}$ from equations \eqref{eqn.stateAprime} and \eqref{eqn.dynamicmodel2} are available. Since the configuration matrix is computed from various estimates with several simplifying assumptions, there inevitably exists some degree of uncertainty. As a result, this configuration matrix only serves as an initial estimate, $\hat{\bm{A}}$, to be further refined by the flight controller.

\section{Adaptive Geometric Flight Controller} \label{sec.controller}

After the flight modules are incorporated with the payload, the IMU-based estimation strategies proposed in Section \ref{sec.imu-based_estimation} are employed to compute the initial estimate of the configuration matrix, $\hat{\bm{A}}$, for flight control purposes. While the estimation error, $\tilde{\bm{A}} := \hat{\bm{A}}-\bm{A}$, may be sufficiently small for the robot to achieve a stable flight, it may still lead to an unsatisfactory flight performance, rendering the system unsuitable for practical applications. In this section, we devise an adaptive geometric flight controller to address the issue.

\subsection{Controller design}

The objective of the flight controller is to ensure that the robot follows a prescribed trajectory. In other words, the position of the robot $\bm{P}$ converges to the setpoint $\bm{P}_\text{d}(t)$. In addition, we attempt to control the yaw orientation of the robot (defined as $\psi$), such that $\psi\rightarrow\psi_\text{d}$, as $\bm{P}$ and $\psi$ constitute the flat outputs of the system. \cite{mellinger2011minimum,mahony2012multirotor,morrell2018differential}.

To control the position, taking motivation from a sliding mode control method and our previous works \cite{chirarattananon2016perching,mu2017adaptive}, we define an error vector $\bm{e}$ that captures the position error and its higher order derivatives:
\begin{equation}
\begin{split}
\bm{e}=&\bm{K}_3(\bm{P}^{(3)}-\bm{P}^{(3)}_\text{d})+\bm{K}_2(\ddot{\bm{P}}-\ddot{\bm{P}}_\text{d})\\
&+\bm{K}_1(\dot{\bm{P}}-\dot{\bm{P}}_\text{d})+\bm{K}_0(\bm{P}-\bm{P}_\text{d})\\
=&\bm{K}_3\bm{P}^{(3)}-\bm{P}_\text{r},
\end{split}\label{eqn.variablee}
\end{equation}
where $\bm{P}^{(3)}$ is the third order time derivative of $\bm{P}$ and $\bm{P}_\text{r}$ is defined accordingly.
Here, $\bm{K}_i$'s are some diagonal matrices with positive elements satisfying $\bm{K}_3=\text{diag} \left(1,1,0 \right)$, $\bm{K}_2=\text{diag} \left(k,k,0 \right)$, $\bm{K}_1=\text{diag} \left(kk_\text{d}, kk_\text{d},1 \right)$, and $\bm{K}_0=\text{diag} \left(kk_\text{p}, kk_\text{p},k_\text{z} \right)$. The reason for having two different sets of gains stems from the inherent dynamics of the robot---the translational dynamics are of fourth order whereas the altitude dynamics are of second order.

According to the Routh-Hurwitz stability criterion, it is guaranteed that $\bm{P}\rightarrow \bm{P}_\text{d}$ when  $\bm{e}\rightarrow \bm{0}$ if the gains satisfy the condition: $k>k_\text{p}/k_\text{d}$ . 

To ensure that $\bm{e}\rightarrow \bm{0}$ and the yaw error is minimized, we construct a composite vector $\bm{s}$ from the projection of $\bm{e}$ onto the body axes and the yaw error and attempt to minimize $\bm{s}$,
\begin{equation}
\bm{s}
=
\left[
  \begin{array}{c}
    \bm{e}\cdot\hat{\bm{k}} \\
    -\bm{e}\cdot\hat{\bm{j}}/g \\
    \bm{e}\cdot\hat{\bm{i}}/g \\
    \bm{\omega}_z+k_\psi(\psi-\psi_\text{d}) \\
  \end{array}
\right],\label{eqn.variables}
\end{equation}
where $k_\psi$ is a positive gain for the yaw control. 
To compute $\bm{s}$ from equations \eqref{eqn.variablee} and \eqref{eqn.variables}, we employ equation \eqref{eqn.positions} and a simplifying assumption that the robot's total thrust is approximately constant, such that $T/m\approx g$. This yields 
\begin{align}
\ddot{\bm{P}}&\approx g\hat{\bm{k}}+\bm{g},\label{eqn.ddP_proj} \text{\quad and} \\ 
\bm{P}^{(3)}&\approx g(-\bm{\omega}_{x}\hat{\bm{j}}+\bm{\omega}_{y}\hat{\bm{i}}). \label{eqn.dddP_proj}
\end{align}
It follows that we can express $\bm{s}$ as a combination of two terms:
\begin{eqnarray}
\bm{s}&\approx
\left[
  \begin{array}{c}
    \dot{z} \\
    \bm{\omega}_{x} \\
    \bm{\omega}_{y} \\
    \bm{\omega}_{z} \\
  \end{array}
\right]
+
\underbrace{
\left[
  \begin{array}{c}
    \dot{z}_\text{d}+k_\text{z}(z-z_\text{d}) \\
    (\bm{P}_\text{r}\cdot\hat{\bm{j}})/g \\
    -(\bm{P}_\text{r}\cdot\hat{\bm{i}})/g \\
    k_\psi(\psi-\psi_\text{d}) \\
  \end{array}
\right]}_{\bm{h}}
=\left[
  \begin{array}{c}
    \dot{z} \\
    \bm{\omega} \\
  \end{array}
\right]
+\bm{h},\label{eqn.s_simplified}
\end{eqnarray}
where $\bm{\omega}_{x}$ and $\bm{\omega}_{y}$ emerge from the term $\bm{P}^{(3)}$ according to equation \eqref{eqn.dddP_proj} and other terms are lumped into $\bm{h}$.

To design an adaptive controller to stabilize the robot in the presence of parameter uncertainties, we propose a Lyapunov function candidate
\begin{equation}
V=\frac{1}{2}\bm{s}^\text{T}\bm{s}+\frac{1}{2}\texttt{tr}(\tilde{\bm{A}}^\text{T}\bm{\Lambda}^{-1}\tilde{\bm{A}}),\label{eqn.lyapunov}    
\end{equation}
where the diagonal matrix $\bm{\Lambda}=\text{diag}(\lambda_{z},\lambda_\phi,\lambda_\theta,\lambda_\psi)\in\mathbb{R}^{4\times 4}$ is positive definite. Both terms in $V$ are positive definite and radially unbounded. The inclusion of the second term enables us to take the estimation errors into consideration. We consider the time derivative of $V$
\begin{equation}
\begin{split}
\dot{V}
=&\bm{s}^\text{T}\dot{\bm{s}}+\texttt{tr}(\dot{\tilde{\bm{A}}}^\text{T}\bm{\Lambda}^{-1}\tilde{\bm{A}}).
\end{split}
\end{equation}
The goal is to find a stabilizing controller and an adaptive law that render $\dot{V}$ negative definite. To do so, we introduce another positive definite diagonal gain matrix $\bm{K}=\text{diag}(k_\text{z}, k_\phi, k_\theta, k_\psi)\in\mathbb{R}^{4\times 4}$. In addition, we define $\bm{g}_v=[g, 0, 0, 0]^\text{T}$, and use equations \eqref{eqn.dynamicmodel2} and \eqref{eqn.s_simplified} to write $\dot{\bm{s}}$ as $\dot{\bm{s}}=\bm{A}\bm{u}-\bm{g}_v+\dot{\bm{h}}=\hat{\bm{A}}\bm{u}-\tilde{\bm{A}}\bm{u}-\bm{g}_v+\dot{\bm{h}}$. The time derivative of $V$ becomes
\begin{equation}
\begin{split}
\dot{V}=&-\bm{s}^\text{T}\bm{K}\bm{s}+\bm{s}^\text{T}\left ( \dot{\bm{s}}+\bm{K}\bm{s} \right ) +\texttt{tr}(\dot{\tilde{\bm{A}}}^\text{T}\bm{\Lambda}^{-1}\tilde{\bm{A}})\\
=&-\bm{s}^\text{T}\bm{K}\bm{s}+\bm{s}^\text{T}\left ( \hat{\bm{A}}\bm{u}-\bm{g}_v+\dot{\bm{h}}+\bm{K}\bm{s} \right )\\ 
&-\bm{s}^\text{T}\tilde{\bm{A}}\bm{u}+\texttt{tr}(\dot{\tilde{\bm{A}}}^\text{T}\bm{\Lambda}^{-1}\tilde{\bm{A}}). \label{eqn.lyapunov_derivative}
\end{split}
\end{equation}
Consequently, we can ensure that $\dot{V}$ is negative definite:
\begin{eqnarray}
\dot{V}=-\bm{s}^\text{T}\bm{K}\bm{s}\leq 0,\label{eqn.lyapunov_stability}
\end{eqnarray}
and the system is asymptotically stable as long as the following conditions hold
\begin{eqnarray}
\hat{\bm{A}}\bm{u}&=&\bm{g}_v-\dot{\bm{h}}-\bm{K}\bm{s}, \text{\quad and}\\
\bm{s}^\text{T}\tilde{\bm{A}}\bm{u}&=&\texttt{tr}(\dot{\tilde{\bm{A}}}^\text{T}\bm{\Lambda}^{-1}\tilde{\bm{A}}).
\end{eqnarray}
The first condition is satisfied by applying the control input
\begin{eqnarray}
\bm{u}=\hat{\bm{A}}^\text{T}(\hat{\bm{A}}\hat{\bm{A}}^\text{T})^{-1}(\bm{g}_v-\dot{\bm{h}}-\bm{K}\bm{s}),\label{eqn.controlinput}
\end{eqnarray}
whereas the second condition necessitates the update law:
\begin{eqnarray}
\dot{\hat{\bm{A}}}=\bm{\Lambda}\bm{s}\bm{u}^\text{T},\label{eqn.adaptivelaw}
\end{eqnarray}
where $\dot{\hat{\bm{A}}}=\dot{\tilde{\bm{A}}}$.
To apply the proposed control and adaptive laws, one requires the knowledge of $\bm{s}$ and $\dot{\bm{h}}$. According to equations \eqref{eqn.variablee} and \eqref{eqn.variables}, it may seem that measurements of $\bm{P}^{(3)}$ and $\ddot{\bm{P}}$ are needed. In practice, when $\bm{P}^{(3)}$ is projected on the body axes, it is approximated as the angular velocity as given by equation \eqref{eqn.dddP_proj}. Similarly, the projections of $\ddot{\bm{P}}$ are related to the attitude of the robot according to equation \eqref{eqn.ddP_proj} \cite{chirarattananon2016perching}. The feedback required for implementation of these laws is standard measurements commonly used by other flight controllers \cite{mahony2012multirotor,mellinger2012trajectory}.

The negative definite property of $\dot{V}$ in equation \eqref{eqn.lyapunov_stability} implies that the value of the Lyapunov function candidate continuously decreases as long as $\bm{s}$ is non-zero. In other words, the position tracking errors and estimation errors are reduced until the position errors vanish. This, however, does not explicitly guarantee that the estimate $\hat{\bm{A}}$ would converge to $\bm{A}$. The estimate, $\hat{\bm{A}}$, is adapted such that the tracking error is reduced, not necessarily for the estimation error to disappear. This is because $\hat{\bm{A}}\rightarrow\bm{A}$ is not a necessary condition for $\bm{s},\bm{e}\rightarrow \bm{0}$. Furthermore, in theory, no speicial condition is required for the initial estimate of $\bm{A}$. In practice, the initial $\hat{\bm{A}}$ must be sufficiently close to $\bm{A}$ for the robot to liftoff stably.

\subsection{Controller gains and scaling analysis}

We have introduced several control parameters $\bm{K}_i$, $\bm{K}$, and $\bm{\Lambda}$. While different vehicular configuration might call for different optimal parameters, it is highly desirable to have one set of parameters that produces reasonable flight performance across various robot configurations in order to avoid the gain tuning process. We provide a simplified analysis to show that this could be achieved.

First, we focus on the translational and rotational dynamics, the primary equation that governs the dynamics of the system is equation \eqref{eqn.dynamicmodel2}. Since the flight modules are attached to the surface of the payload, if the robot's configuration is assumed to be disc-like, the number of flight modules is approximately proportional to the squared characteristic length of the robot such that $N\sim (\bm{r}^\text{cm}_i)^2$ or $\bm{r}^\text{cm}_i\sim \sqrt{N}$ (a similar assumption is employed in \cite{oung2014distributed}). From the fact that $m\sim N$ as suggested by equation \eqref{eqn.mass_estimation}, it follows that $\bm{I}\sim m \bm{r}^2\sim N^2$, and $\dot{\bm{\omega}}_i\sim N^{-2}\cdot N\sqrt{N} \cdot \bm{u}_i \sim N^{-1/2}\bm{u}_i$.

To understand the consequence of this scaling effect in the context of the closed-loop system according to our Lyapunov analysis, we consider equation \eqref{eqn.lyapunov_derivative} in the ideal condition ($\tilde{\bm{A}}=\bm{0}$), the control law for $\bm{u}_i$ (equation \eqref{eqn.controlinput}) was chosen such that $\bm{A}\bm{u}=\bm{g}_v-\dot{\bm{h}}-\bm{K}\bm{s}$ regardless of $N$. If all control parameters ($\bm{K}_i$, $\bm{K}$, and $\bm{\Lambda}$) are held constant, this means the magnitude of $\bm{u}_i$ is expected to scale as $\sqrt{N}$ to maintain this condition as the vehicle size increases. While this is not ideal, it is proven experimentally feasible for $N=4$ to 8.

To analyze the effects of disturbances, we assume there exists some noise associated with each propelling module. For ease of analysis, this is presented as an extra term in the input such that $\bm{u}_i\rightarrow \bm{u}_i + w$. This $w$ could be a result of unreliable hardware or wind disturbances (the effects of wind on the propellers are more pronounced than on the body due to the aerodynamic interactions \cite{mahony2012multirotor}). In this scenario, $w$ eventually leads to an undesirable term in equation \eqref{eqn.lyapunov_derivative} that is proportional to $\sim \left \| \bm{s} \right \| N^{-1/2} w$. In experiments, this could be considered one factor that prevents the robot from converging to the setpoint. It can be seen that, since this term becomes less significant as $N$ grows, it suggests that the strategy to keep all control parameter constants should result in the reduced position errors in larger robots, at the cost of more demanding control efforts. A similar framework, when apply to the altitude dynamics, reveals that the anticipated altitude error is independent of $N$.

In fact, a more rigorous treatment on the scaling and controller gains can be found in \cite{oung2014distributed}. Therein, the authors computed the controller gains using the H2-optimal control method based on the linearized dynamics of the robots. Though, this necessitates the full knowledge of the robot's configuration and disturbance model. In the implementation across multiple vehicle sizes, the proposed strategy was compared with a constant set of gains. The results show that the constant gains approach, while not optimal, provides satisfactory flight performance, with the errors decrease as the vehicle size grows, consistent with our prediction.

\section{Prototypes} \label{sec.prototypes}

To verify the concept of the modular design and the proposed estimation and control strategies, we manufacture the flight modules and incorporate them with dummy payloads to create six different robots.
\begin{table}[ht]
\begin{center}
\caption{Major components of flight modules.}
\begin{tabular}{lc}\hline\hline%
\textbf{Item Types}   & \textbf{Descriptions} \\ %
\hline %
\multicolumn{2}{c}{\textbf{Propelling module}}\\
\hline %
LiPo battery & Turnigy Bolt 550mAh \\ %
\hdashline[0.1pt/2pt]
Propeller & Quanum Carbon Fiber $6\times4.5$ inch \\ %
\hdashline[0.1pt/2pt]
Motor+ESC & ZTW BW 2204 Series 2300KV \\ %
\hdashline[0.1pt/2pt]
IMU & MPU9250 \\ %
\hline %
\multicolumn{2}{c}{\textbf{Control module}}\\
\hline %
LiPo battery & Turnigy Bolt 550mAh \\ %
\hdashline[0.1pt/2pt]
Autopilot & Pixfalcon \\ %
\hdashline[0.1pt/2pt]
Communication & FrSky X6R Receiver \\%
modules       & XBee-Pro S1 Modules \\%
\hdashline[0.1pt/2pt]
Single-board & \multirow{2}{*}{Raspberry Pi Zero W} \\%
computer & \\%
\hdashline[0.1pt/2pt]
Power module & APM POWER Module with 5.3V DC BEC \\ %
\hline\hline%
\end{tabular}\label{tab.module_component}
\end{center}
\end{table}
\begin{figure*}[ht]
\centering
\psfrag{scale}[c][c][0.9]{2 cm}
\psfrag{motor}[c][c][0.9]{motor}
\psfrag{pro}[c][c][0.9]{propeller}
\psfrag{bat}[c][c][0.9]{battery}
\psfrag{mount}[c][c][0.9]{~~~~~~~~adhesive pad}
\psfrag{sbc}[c][c][0.9]{~~~~~~~~~~~~~~~~~~~~~~~~~~~single-board computer}
\psfrag{PCB}[c][c][0.9]{PCB board}
\psfrag{auto}[c][c][0.9]{autopilot}
\psfrag{com}[c][c][0.9]{communication modules}
\psfrag{power}[c][c][0.9]{power module}
\psfrag{IMU}[c][c][0.9]{IMU}
\psfrag{x}[c][c][0.9]{X}
\psfrag{y}[c][c][0.9]{Y}
\psfrag{z}[c][c][0.9]{Z}
\psfrag{cm}[c][c][0.9]{cm}
\subfigure[a propelling module]{\includegraphics[width=80mm]{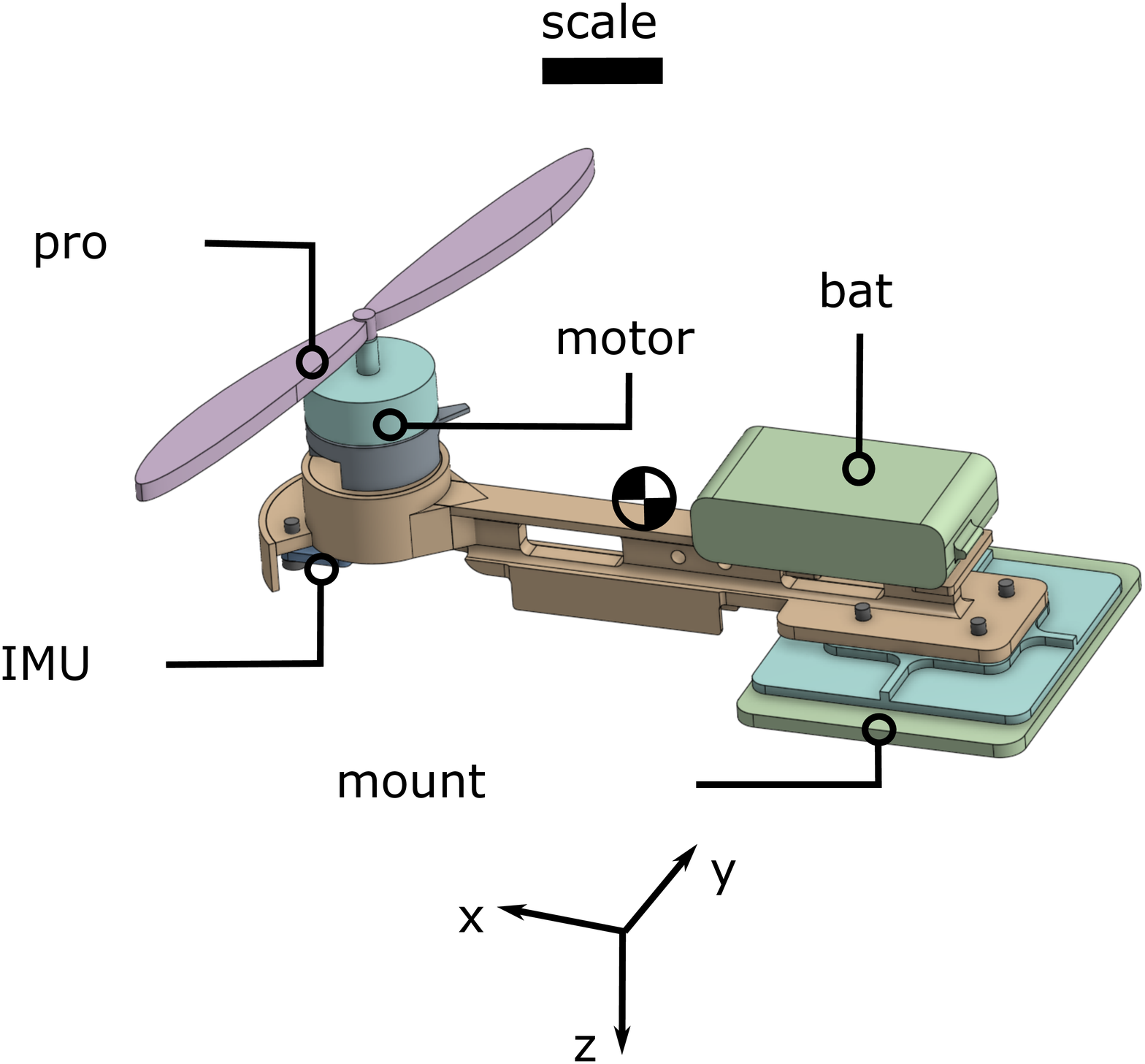}}
\subfigure[a control module]{\includegraphics[width=80mm]{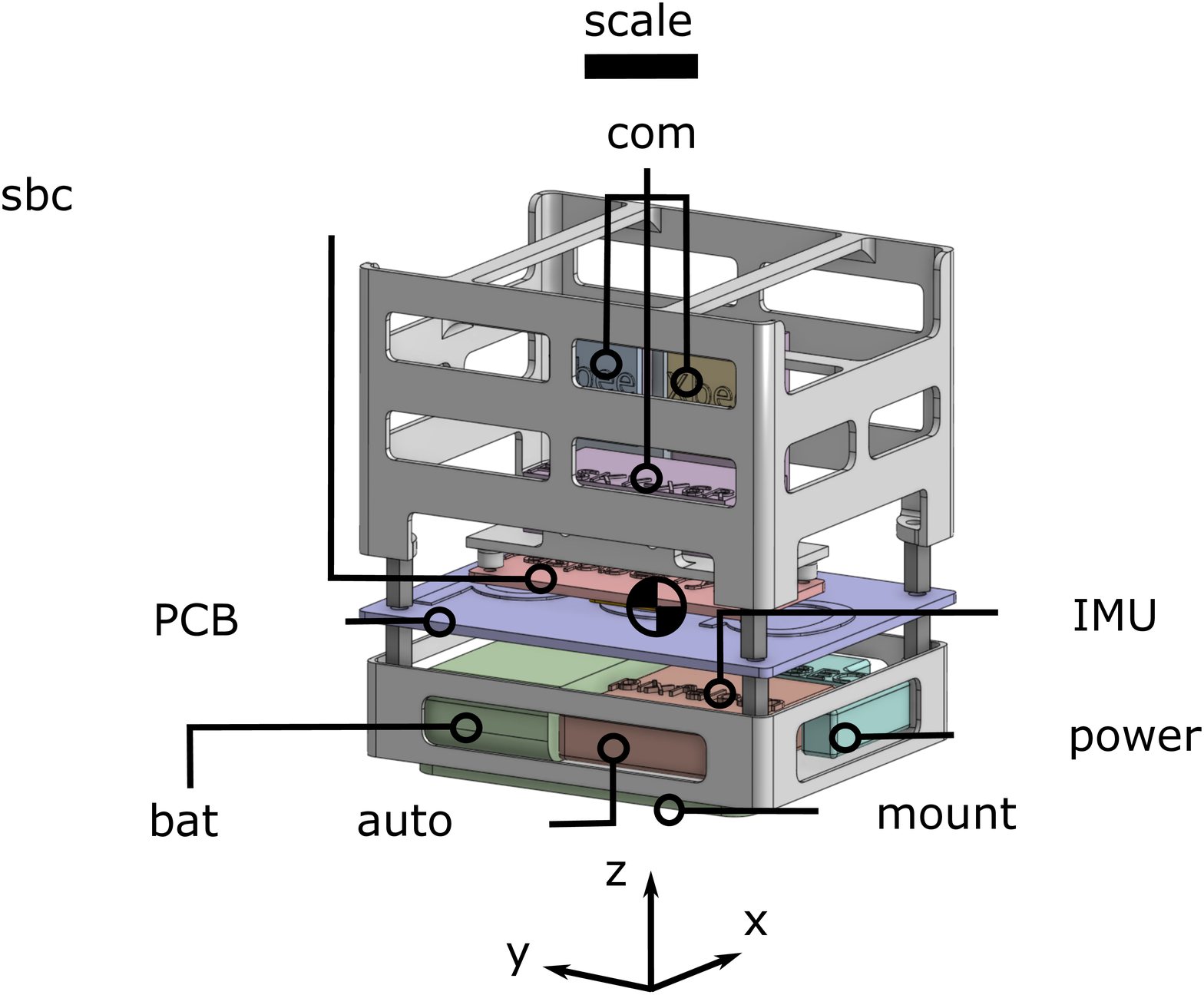}}
\caption{CAD drawings of (a) a propelling module and (b) a control module. The module's frame is defined to be coincident with the IMU's frame.}\label{fig.modulesCAD}
\end{figure*}

\subsection{Flight modules}

Major components of a propelling module are listed in Table \ref{tab.module_component}. As shown in Fig. \ref{fig.modulesCAD}(a), each propelling module has a brushless motor with a built-in electronic speed controller (ESC) and a 6-inch propeller. The IMU required for the estimation is located directly below the propeller, simplifying the calculation. Instead of sharing one power source for the entire robot, each module individually carries a 550mAh battery. This design provides more consistent flight endurance for robots with a different number of propelling modules. Local communication with the control module is wired. The parts are held together by an elongated 3D-printed structure with the length of 16.5 cm. Each propelling module weighs 130 g. The propeller was measured for generated thrust and torque when subject to different commands using a multi-axis load cell (ATI nano 25) on a setup similar to \cite{hsiao2018ceiling}. This allows $T_i$'s in equation \eqref{eqn.stateAprime} to be directly used as the system's inputs. The ratio between the propeller's torque to thrust ($c_i$ in equation \eqref{eqn.stateAprime}) is directly obtained without the need to measure the spinning rate. The maximum thrust from one module is 6.5 N and $c_i=\pm 0.016$. Other physical parameters are provided in Table \ref{tab.modules}. Therein, the inertia tensor of the flight module is computed with respect to its own CM location. 
\begin{table*}[ht]
\begin{center}
\caption{Some important properties of the propelling and control modules.}
\small{
\begin{tabular}{lrrrrrr}\hline\hline%
\textbf{Properties} & \multicolumn{3}{c}{\textbf{Propelling modules}} & \multicolumn{3}{c}{\textbf{Control module}} \\ \cline{1-1} \cline{2-4} \cline{5-7} %
\textbf{Size} (cm) & \multicolumn{3}{c}{[24.3,~~15.6,~~6.0]} & \multicolumn{3}{c}{[7.4,~~9.4,~~9.9]}\\ \hdashline[0.1pt/2pt]
$m_i$ (g) & \multicolumn{3}{c}{130} & \multicolumn{3}{c}{250}\\ \hdashline[0.1pt/2pt]
\textbf{Inertia tensor}
& $5.02\times10^{-5}$ & $-1.22\times 10^{-7}$ & $3.17\times 10^{-5}$ & $2.76\times 10^{-4}$ & $-3.86\times 10^{-7}$ & $-1.59\times 10^{-5}$\\ %
(kg$\cdot$m$^2$) & $-1.22\times 10^{-7}$ & $3.85\times 10^{-4}$ & $8.15\times 10^{-8}$ & $-3.86\times 10^{-7}$ & $2.39\times 10^{-4}$ & $-1.18\times 10^{-5}$\\ %
& $3.17\times 10^{-5}$ & $8.15\times 10^{-8}$ & $3.86\times 10^{-4}$ & $-1.59\times 10^{-5}$ & $-1.18\times 10^{-5}$ & $2.32\times 10^{-4}$\\ \hdashline[0.1pt/2pt]
$T_i$ (N) & \multicolumn{3}{c}{[0,~~6.5]} & \multicolumn{3}{c}{-}\\  \hdashline[0.1pt/2pt]
$c_i$ (m) & \multicolumn{3}{c}{$\pm1.6\times 10^{-2}$} & \multicolumn{3}{c}{-} \\
\hline\hline%
\end{tabular}}\label{tab.modules}
\end{center}
\end{table*}

The control module, illustrated in Fig. \ref{fig.modulesCAD}(b), primarily consists of a flight control unit (with a built-in IMU), communication modules and a battery. As listed in Table \ref{tab.module_component}, we employ a Pixfalcon autopilot for implementation of the customized flight controller through Simulink (MathWorks) \cite{PX42016}. XBees and a telemetry receiver are used for communication with the ground computer. A single-board computer, Raspberry Pi, are connected with all the IMUs for the parameter estimations. A custom PCB board is manufactured for reliable electrical connections. All components are housed together and protected from collisions with a 3D-printed structure. The total mass of the control module is 250 g.

\begin{figure*}[ht]
\centering
\psfrag{cm}[c][c][0.9]{10 cm}
\psfrag{pa}[c][c][0.9]{~~payload: 208 g}
\psfrag{pb}[c][c][0.9]{~payload: ~~235 g}
\psfrag{pc}[c][c][0.9]{~~payload: ~~361 g}
\psfrag{pd}[c][c][0.9]{~~~~~~~~payload: 228 g}
\psfrag{pe}[c][c][0.9]{~~~~~~payload: ~317 g}
\psfrag{pf}[c][c][0.9]{~~~~~~~~~~payload: ~791 g}
\psfrag{ta}[c][c][0.9]{~~~~~~total: 978 g}
\psfrag{tb}[c][c][0.9]{~~~~~~total: 1265 g}
\psfrag{tc}[c][c][0.9]{~~~~~~~total: 1651 g}
\psfrag{td}[c][c][0.9]{~~~~~~~~~~~~total: 998 g}
\psfrag{te}[c][c][0.9]{~~~~~~~~~~total: 1347 g}
\psfrag{tf}[c][c][0.9]{~~~~~~~~~~~~~~total: 2081 g}
\psfrag{x}[c][c][0.9]{X}
\psfrag{y}[c][c][0.9]{Y}
\psfrag{A}[c][c][0.9]{~~~~~~~~~~~~~Platform A}
\psfrag{B}[c][c][0.9]{~~~~~~~~~~~~~Platform B}
\psfrag{C}[c][c][0.9]{~~~~~~~~~~~~~Platform C}
\psfrag{D}[c][c][0.9]{~~~~~~~~~~~~~Platform D}
\psfrag{E}[c][c][0.9]{~~~~~~~~~~~~~Platform E}
\psfrag{F}[c][c][0.9]{~~~~~~~~~~~~~Platform F}
\includegraphics[width=170mm]{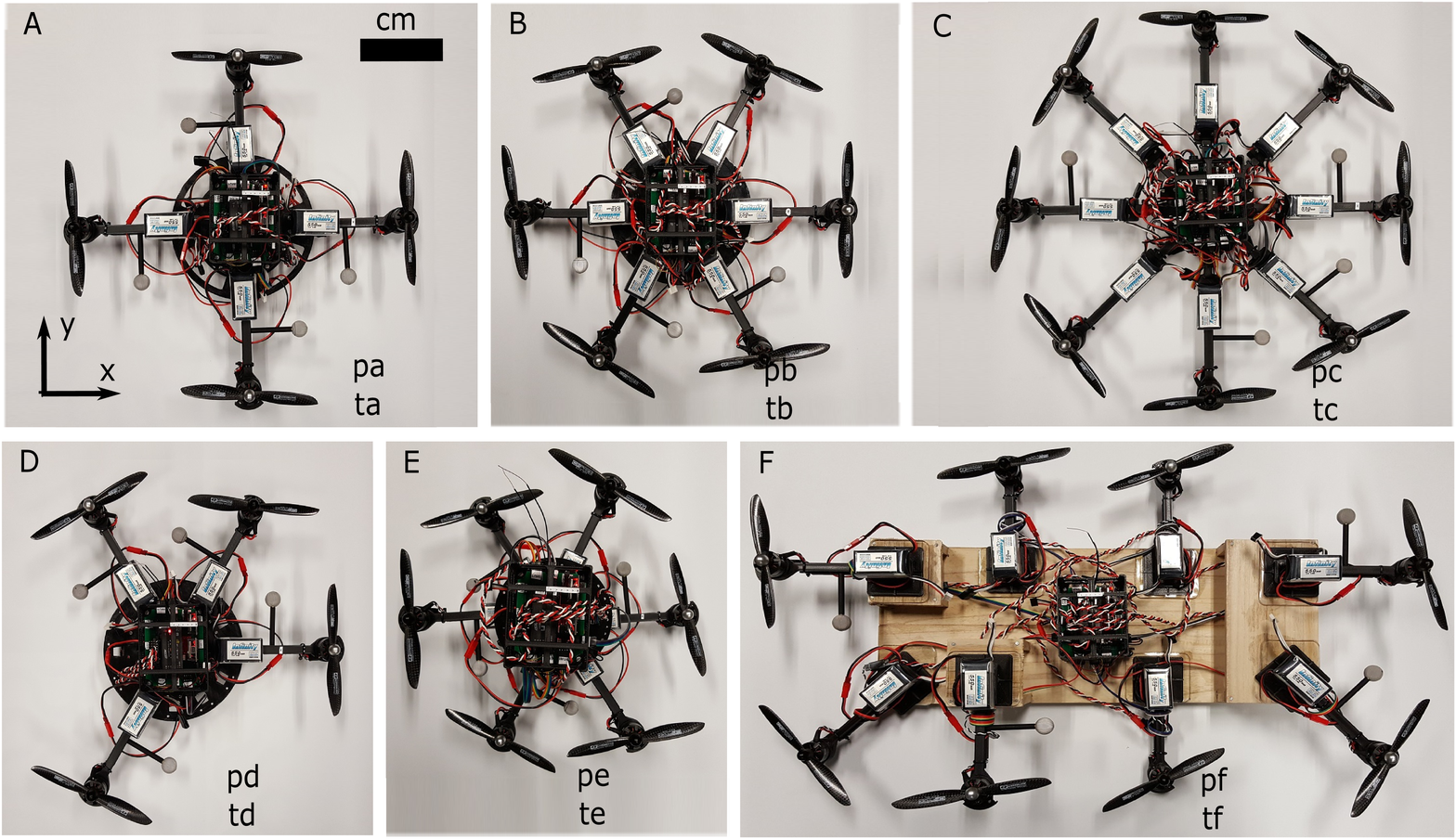}
\caption{Six prototypes shown in the horizontal plane. The propelling modules of platforms A, B, C are symmetrically placed, and the propelling modules of platforms D, E, F are placed in asymmetrical configurations. The propelling modules of platforms A, B, C, D are at the same height, whereas the propelling modules of platforms E, F are located at different heights.}\label{fig.prototypes}
\end{figure*}

\subsection{Modular robot prototypes}
To demonstrate the concept of modular robot design, where flight modules can be incorporated with different payloads in various configurations, we constructed six different platforms as shown in Fig. \ref{fig.prototypes}. In each prototype, as limited by our assumptions, the payload is rigid and the thrust directions are approximately aligned. 

To verify the scalability of the modular design, the constructed prototypes have 4, 6, and 8 propelling modules. For comparison, platforms A, B, and C have symmetric configurations similar to conventional multirotor MAVs, whereas platforms D, E, and F have irregular arrangements of propelling modules. In addition, the propelling modules of platforms E and F are attached to the payload at different heights.

For convenience, we manufactured different disc-like dummy payloads via 3D-printing for prototypes A-E. The masses and sizes (diameter of the planar discs) of these payloads vary from $\approx200-360$ g and $13-16$ cm as given in Fig. \ref{fig.prototypes}. The control modules were placed near the center of the payloads. The propelling modules were attached around the edge of the printed payloads using screws. To simulate a real-world use, platform F was made from a $50\times20\times11$ cm piece of wood with areas for attaching the propeller modules located at different heights. The payload mass is 791 g. Each flight module was attached to the payload via a $7.6\times6.2$ cm commercial off-the-shelf reusable gel pad adhesive instead of screws. This adhesive option was chosen for demonstration owing to the ease of use and can be substituted by other attachment mechanisms. According to the manufacturer (Stikk Gel Pads), each gel pad holds up to $2$ kg and can be re-used over 200 times after washing. The strength of the double-sided adhesive may be reduced when used with rough surfaces, wet or porous materials. This imposes some limitation on the suitable payloads, but the pads can be easily replaced in case of deteriorated adhesion.

\section{Pre-flight Estimations} \label{sec.prototypeparamters}
\subsection{Mass estimates}

The total mass of each prototype is estimated according to equation \eqref{eqn.mass_estimation}. The mass of the flight modules ($m_i$) are directly taken from Table \ref{tab.modules}. We used $m_n=50$ g, such that each propelling module is  responsible for carrying approximately 50 g of payload. Across platforms A-E, this results in the RMS errors of 37 g. For platform F, the mass estimate underpredicts the payload by $\approx400$ g. In all prototypes, the payloads account for at least $20\%$ of the actual weight of the vehicles. For further details of mass, refer to Table \ref{tab.Mass} in the Supplemental Materials.

\subsection{Configuration matrix estimates}
To obtain the estimate of the configuration matrix using the IMU-based method as proposed in Section \ref{sec.imu-based_estimation}, each prototype was handheld and manually rotated. We ensured the rotation about all axes were achieved. In each trial, we carried out the rotation for over 90 seconds, resulting in over 200 measurements from each IMU. The measurements are time-synchronized on the onboard Raspberry Pi. We repeated the process five times for each prototype.

The gyroscopic measurements were low-pass filtered to eliminate noises (cutoff frequency: $4$ Hz). We applied the least-squares method according to equation \eqref{eqn.modulerationmatrix} to compute $\bm{R}_i$, the orientation of the propelling modules with respect to the control module (body frame). The numerical results are normalized to ensure the special orthogonal condition of the rotation matrix. Then, together with the accelerometer readings, we evaluated $\bm{r}_i^\text{gc}$ using equation \eqref{eqn.imualgorithmp}.
\begin{figure*}[ht!]
\centering
\psfrag{PA}[c][c][0.9]{Platform A}
\psfrag{PB}[c][c][0.9]{Platform B}
\psfrag{PC}[c][c][0.9]{Platform C}
\psfrag{PD}[c][c][0.9]{Platform D}
\psfrag{PE}[c][c][0.9]{Platform E}
\psfrag{PF}[c][c][0.9]{Platform F}
\psfrag{ar}[c][c][0.85]{1.6 cm}
\psfrag{aa}[c][c][0.85]{5.2$^\circ$}
\psfrag{br}[c][c][0.85]{2.3 cm}
\psfrag{ba}[c][c][0.85]{8.9$^\circ$}
\psfrag{cr}[c][c][0.85]{2.8 cm}
\psfrag{ca}[c][c][0.85]{8.1$^\circ$}
\psfrag{dr}[c][c][0.85]{2.5 cm}
\psfrag{da}[c][c][0.85]{7.6$^\circ$}
\psfrag{er}[c][c][0.85]{2.4 cm}
\psfrag{ea}[c][c][0.85]{8.7$^\circ$}
\psfrag{fr}[c][c][0.85]{3.0 cm}
\psfrag{fa}[c][c][0.85]{8.5$^\circ$}
\psfrag{x}[c][c][0.9]{X (cm)}
\psfrag{y}[c][c][0.9]{Y (cm)}
\psfrag{Center}[c][c][0.9]{CM~~~~}
\psfrag{Actual}[c][c][0.9]{Actual}
\psfrag{Estimates}[c][c][0.9]{Estimate~}
\includegraphics[width=190mm]{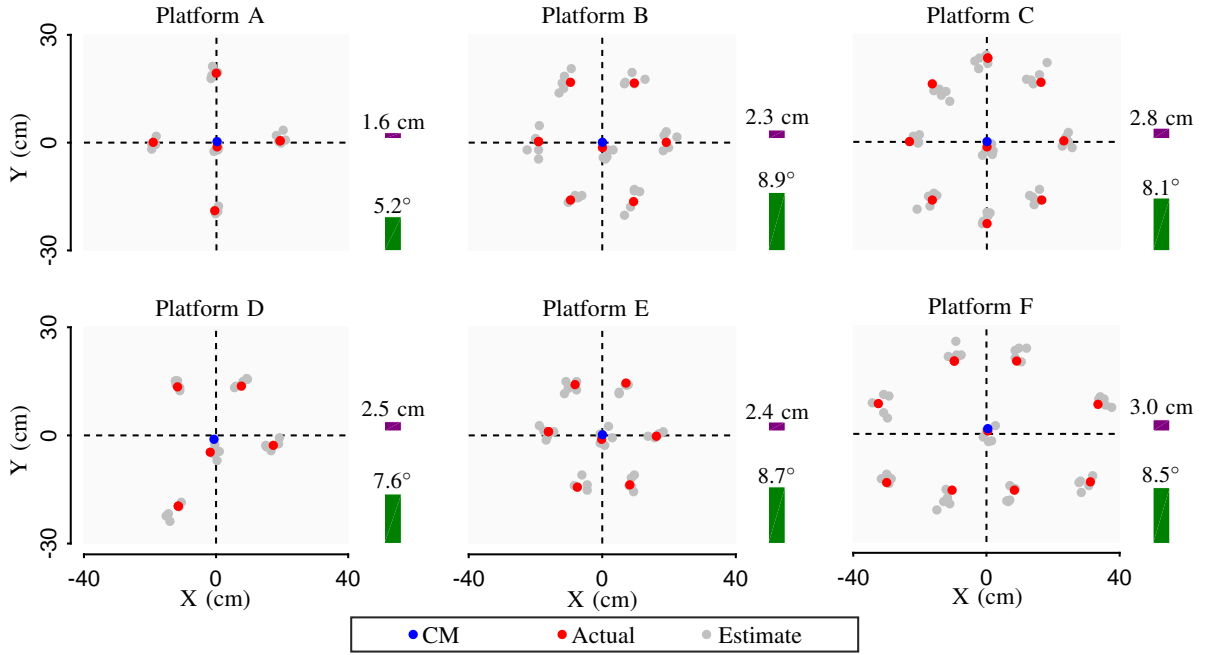}
\caption{The estimates of the locations of all IMUs in the horizontal plane. In each plot, the GC  of all $N+1$ IMUs is at the origin. The blue dots denote the true CM locations. The red dots denote the groundtruth locations of the IMUs and the gray dots denote their corresponding estimates. The side bars show the RMSEs of the estimated IMU locations in terms of distance and angle.}\label{fig.imu_shapes}
\end{figure*}
\begin{figure}[ht!]
\centering
\psfrag{PA}[c][c][0.9]{Platform A}
\psfrag{PB}[c][c][0.9]{Platform B}
\psfrag{PC}[c][c][0.9]{Platform C}
\psfrag{PD}[c][c][0.9]{Platform D}
\psfrag{PE}[c][c][0.9]{Platform E}
\psfrag{PF}[c][c][0.9]{Platform F}
\psfrag{IF}[c][c][0.9]{$\bm{I}_f$}
\psfrag{IT}[c][c][0.9]{$\bm{I}_t$}
\psfrag{AF}[c][c][0.9]{$\bm{A}_f$}
\psfrag{AT}[c][c][0.9]{$\bm{A}_t$}
\psfrag{M}[c][c][0.9]{$m$}
\psfrag{L}[c][c][0.9]{$L$}
\psfrag{N}[c][c][0.9]{$N$}
\includegraphics[width=85mm]{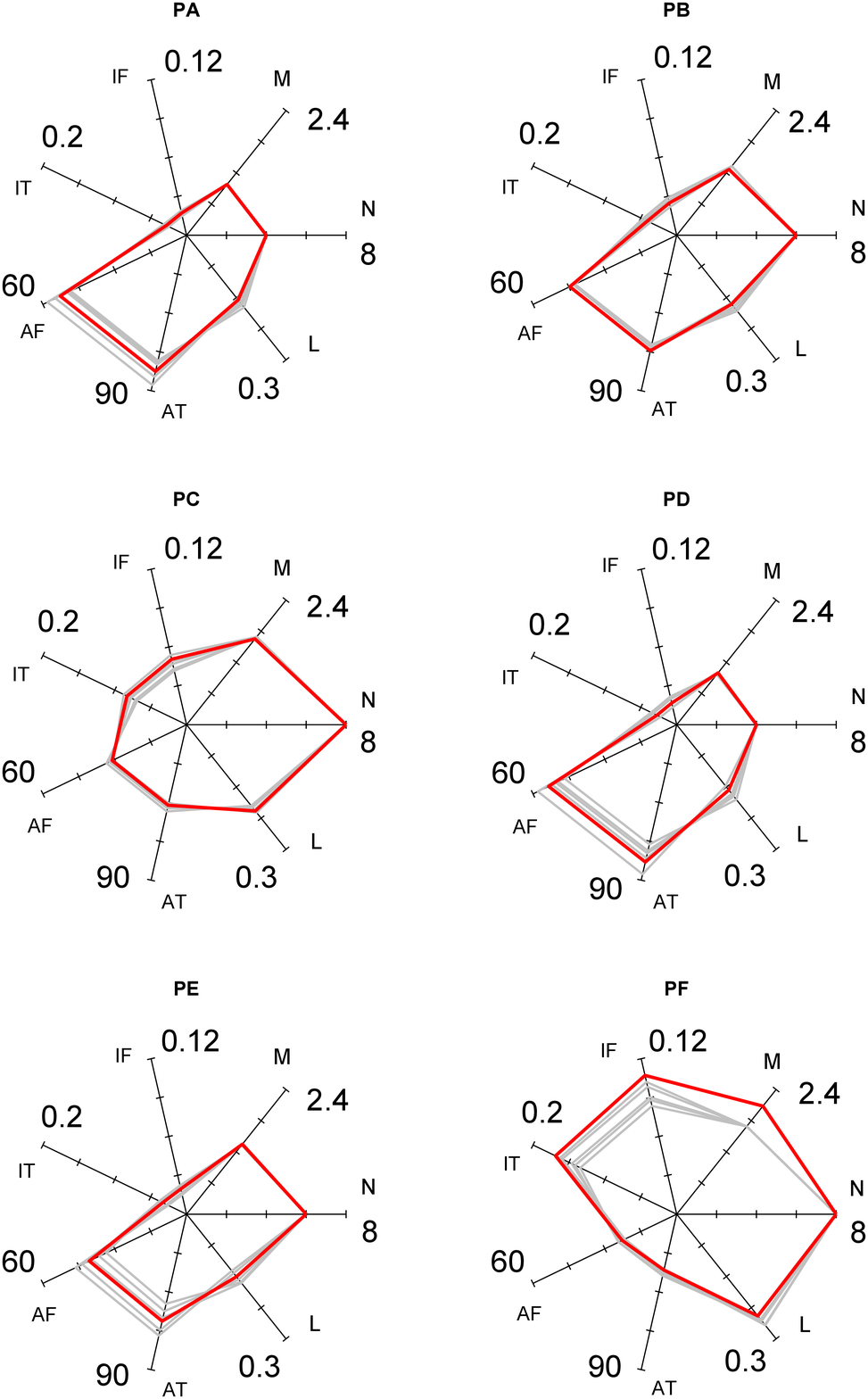}
\caption{The resultant estimates of representative parameters for all platforms. The subscripts `f' and `t' of the inertia tensor $\bm{I}$ and $\bm{A}$ denote the matrix's Frobenius norm and trace norm. The prototype's lengthscale ($L$) is defined as the averaged length of all $\bm{r}^\text{cm}_{i}$. Note that, the units of $m$, $\bm{I}$, and $L$ are kg, kg$\cdot$m$^2$ and m.}\label{fig.deltaerror}
\end{figure}

The estimation results are illustrated in Fig. \ref{fig.imu_shapes}. The GC is placed at the origin of each plot. The actual CM and locations of flight modules from the CAD models are shown for reference. The estimates of $\bm{r}_i^\text{gc}$ are represented as the modules' positions with respect to the GC. The quality of the estimates is assessed as the RMSEs of the magnitude of $\bm{r}_i^\text{gc}$ and RSMEs of the orientation of $\bm{r}_i^\text{gc}$ (measured about the Z axis) shown as side bars in Fig. \ref{fig.imu_shapes}. In all prototypes and all trials, the RMSEs of the modules' locations vary from 1.6 to 3.0 cm while the RMS of the angular errors measured are below 9$^\circ$ across six prototypes.

With the estimates of $m_p$, $\bm{R}_i$, and $\bm{r}_i^\text{gc}$, we employed the strategy in Section  \ref{sec.est_inertia} to estimate the moments of inertia of the prototypes. The results, alongside the CAD estimates, are provided in Table \ref{tab.prototypes_inertia} in the Supplemental Materials. From here, the estimates of configuration matrix $\hat{\bm{A}}$ were obtained. These serve as initial estimates for flight experiments. The estimation results and the corresponding values from the CAD models can be found in Tables \ref{tab.prototypes_A0} in the Supplementary Materials.

To further provide some visual indication of the accuracy of the estimates, Fig. \ref{fig.deltaerror} shows the resultant estimates of $\bm{I}$ and $\bm{A}$ of all six prototypes from five trials with the groundtruth value. Since these are matrices, we use the Frobenius norm (Schatten 2-norm) and trace norm (Schatten 1-norm) for comparison. The Frobenius norm is invariant under rotations. Similarly, the trace norm, for an inertia tensor, is a sum of the inertia along the principal axes, invariant of the change of basis. For these reasons, they are reasonable quantities for comparison. Other important parameters, such as the vehicle's mass and lengthscale are included. It can be seen that, across all six prototypes, these values differ significantly, but all estimates are reasonably accurate. This testifies that our estimation strategies are valid across the relevant scales.

\section{Flight Experiments} \label{sec.experiments}

To demonstrate flights of the proposed modular vehicles, we carried out hovering flights using all six prototypes and trajectory following flights with Platforms A and E, with the estimated configuration matrices from Section \ref{sec.prototypeparamters} as the initial estimates for $\hat{\bm{A}}$.

\subsection{Experimental setup}

The flights were conducted in an indoor flight arena equipped with the motion capture system (NaturalPoint, OptiTrack), covering a volume of $3.6\times3.6\times2.5$ meter. The cameras track the retroreflective markers on the robots for position and attitude feedback. The motion capture system is solely for the groundtruth measurements and real-time position and yaw angle control as commonly found in literature \cite{chirarattananon2016perching,zhao2018design}. The position and yaw angle are wirelessly transmitted from the ground computer running the Simulink Real-Time  environment (Mathworks) to the robots at the rate of $\approx75$ Hz via XBee modules. There is $\approx 100-300$ ms latency in the wireless communication system. The ground station records flight trajectories and essential debug data received from the robots. The attitude control relies entirely on the onboard feedback from the single IMU residing on the control module. The remote control (RC) is used to initiate and terminate the flight, and it is also to provide the signals used for trimming.

The adaptive geometric controller is implemented on onboard autopilot, executing at a rate of 150 Hz. The configuration matrix is adaptively updated at a lower rate of 25 Hz to reduce the computational load. A filter for estimating the position of the robot (refer to \cite{goodarzi2013geometric} for an example) is not implemented. As a result, intermittent data loss in the wireless transmission may occasionally cause the robot to appear relatively unsteady during flight. 

\subsection{Hovering flights} \label{sec.exp.hovering}
\begin{figure*}[ht]
\psfrag{Xp}[c][c][0.9]{X position (cm)}
\psfrag{Yp}[c][c][0.9]{Y position (cm)}
\psfrag{Zp}[c][c][0.9]{Z position (cm)}
\psfrag{time}[c][c][0.9]{time (s)}
\centering
\subfigure[Platform E]{\includegraphics[width=85mm]{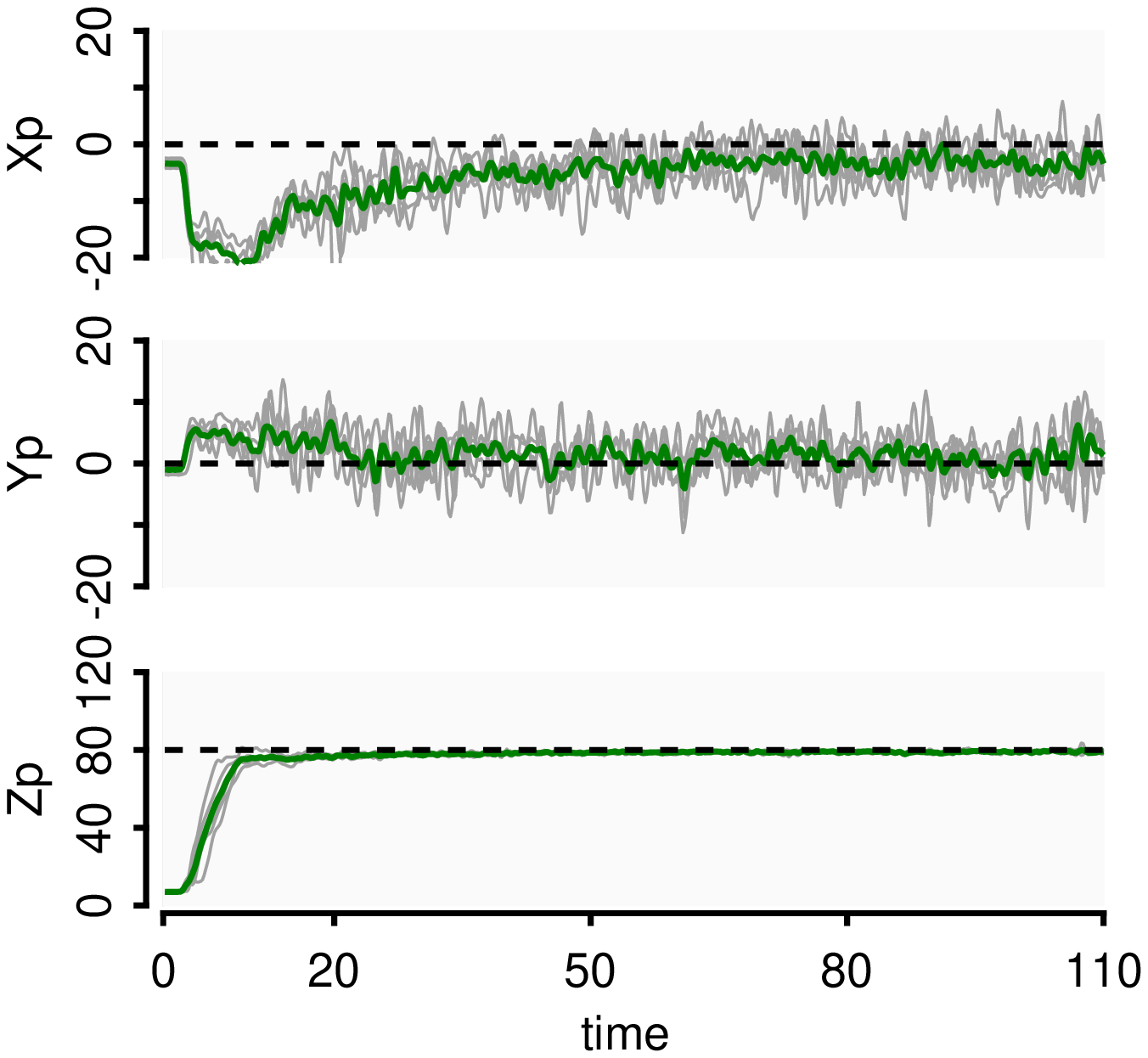}}
\subfigure[Platform F]{\includegraphics[width=85mm]{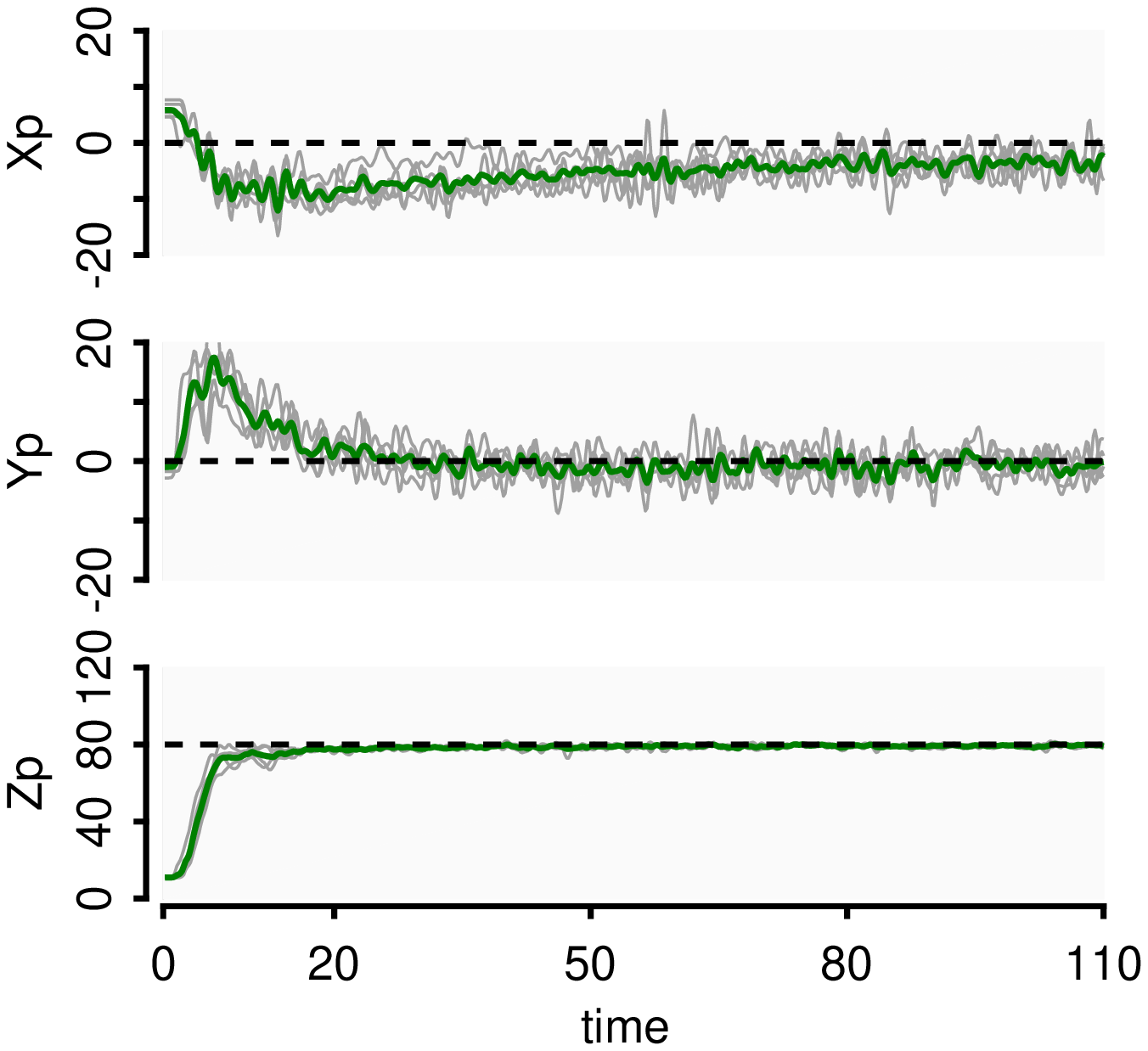}}
\caption{The trajectories of platforms (a) E and (b) F during five hovering flights (gray lines)  with respect to the setpoint (black dashed lines). The dark green lines are the averages from five flights.}\label{fig.hoveringflights110_trajectory_ef}
\end{figure*}
\begin{figure}[ht]
\psfrag{Xe}[c][c][0.9]{X error (cm)}
\psfrag{Ye}[c][c][0.9]{Y error (cm)}
\psfrag{Ze}[c][c][0.9]{Z error (cm)}
\centering
\includegraphics[width=85mm]{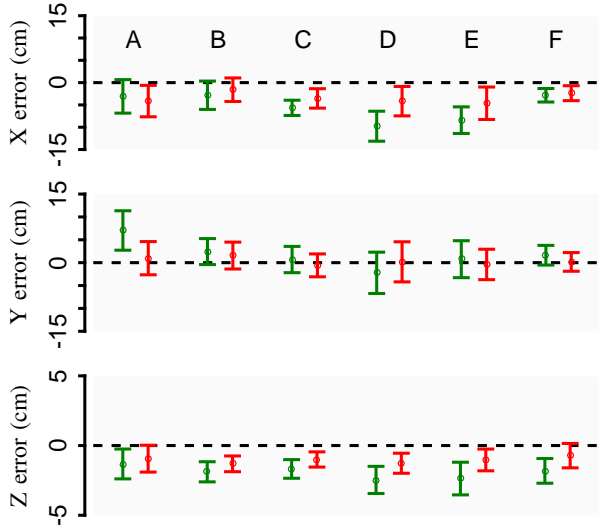}
\caption{The average and standard deviation of the robot's position errors with respect to the setpoint. The green markers are calculated from the time intervals 20 s $\leq t \leq$ 50 s and the red markers are taken from 50 s $\leq t \leq$ 110 s. The plot illustrates the notable decrease in the position errors over time thanks to the adaptive method. }\label{fig.hoveringflights110}
\end{figure}
\begin{figure}[t]
\psfrag{Vs}[c][c][0.95]{$V_s$}
\psfrag{time}[c][c][0.9]{time (s)}
\centering
\includegraphics[width=85mm]{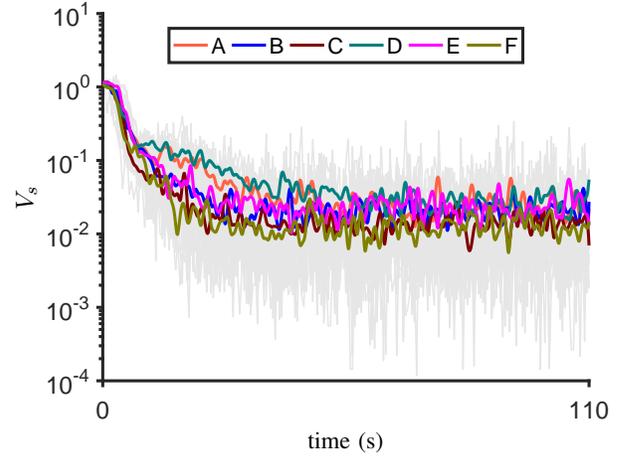}
\caption{The partial Lyapunov function $V_s=\frac{1}{2}\bm{s}^\text{T}\bm{s}$ versus time. The solid lines are the low-pass filtered values from six platforms (averaged from five fligths for each platform). The gray lines present the values from all 30 flights.}\label{fig.stateV}
\end{figure}

\subsubsection{Takeoff flights with no adaptive control}

First, to verify whether the initial estimates of the configuration matrix obtained from IMU-based estimation strategy are sufficiently accurate for the robots to achieve stable flight, we disabled the adaptive algorithm in the flight controller by setting the adaptive gain to zero ($\bm{\Lambda}=\bm{0}$). For platform A, which has a symmetrical configuration, we performed three test flights. In all flights, the robot successfully lifted off. However, the controller was unable to minimize the position errors and keep the robot in the flight arena. In all flights, the robot resulted in crashes within 30 s. The outcomes suggest that the initial estimates, in this case, was sufficient for the robot to attain some degree of attitude stability, but the flight performance was inadequate for practical uses. When the same conditions were applied to platform D, the vehicle failed to take off. This is likely due to the highly imbalanced torques resulted from the asymmetric configuration of platform D. Even with the accurate model of the plant, it is perceivable more difficult to stabilize platform D. To overcome this, we applied some trimming according to the procedure described in section \ref{sec.trimming} in Supplemental Materials. The trimming process modifies the estimate of the configuration matrix according to the user's input. The vehicle successfully lifted off thereafter. However, in all four attempts, the robot flew out of the flight arena in less than 10 s. The test results indicate that the estimation scheme and trimming process may be sufficient to achieve flight, but inadequate to ensure satisfactory flight performance. Further takeoff tests with other platforms reveal that trimming is only required for platforms D and F.

\subsubsection{Sustained flights with the adaptive geometric controller}

Here, we enabled the adaptive part of the proposed geometric flight controller. For each prototype, we performed five hovering flights with the desired position setpoint $\bm{P}_\text{d}=[0,0,80]^\text{T}$ cm, for the duration of 120 s (with the last 10 s reserved for landing). This amounts to 30 flights across six platforms. The same controller gains (refer to Table \ref{tab.parameters_tuning} in the Supplemental Materials) were used for all prototypes. Note that, only platforms D and F were required to be trimmed for stable takeoffs.

All prototypes demonstrated stable flights and stayed within the 3.6$\times$3.6 m flight arena for the entire period. Fig. \ref{fig.hoveringflights110_trajectory_ef} illustrates the trajectories of all five flights belonging to platforms E and F. It can be seen that, within the first 30-40 s, the position errors are radically reduced, and the robots hovered near the setpoint for the rest of the flights. The position errors are approximately 5 cm or less in all directions. Similar results of other prototypes are shown in Fig. \ref{fig.hoveringflights110_trajectory_abcd} in Supplemental Materials. The detailed information of the flight video of all prototypes are described in Table \ref{tab.vedios} in the Supplemental Materials. 

To quantify the performance of the adaptive controller, we plot the average position errors of all platforms with respect to the setpoint in Fig. \ref{fig.hoveringflights110} (30 flights in total). To highlight the contribution of the adaptive algorithm, we consider two separate time intervals: 20 s $\leq t \leq$ 50 s and 50 s $\leq t \leq$ 110 s. The former interval presents the portion of stable flight (long enough after to be affected by the transients from the takeoff maneuvers), with the uncertain parameters still adapting. In the latter part, the tracking errors do not vary significantly over time. This indicates that the estimates of the configuration matrices have, to large extent, converged. It can be seen from Fig. \ref{fig.hoveringflights110} that, in the latter interval, the robots achieved relatively stable flight, with the average position errors lower than 4.7 cm, 1.6 cm and 1.4 cm in x-axis, y-axis and z-axis. These are notably lower than the average errors from the first intervals (9.8 cm, 7.0 cm, and 2.5 cm). The results verify that the adaptive algorithm radically improves the flight performance. After the parameter convergence, the tracking errors of our platforms are comparable to those of similar robots with known robot parameters \cite{oung2014distributed,zhao2018design}.

Furthermore, the flight performance can also be quantified in terms of the Lyapunov function. According to equations \eqref{eqn.lyapunov} and \eqref{eqn.lyapunov_stability}, the controller adaptively decreases the value of $V$ as long as the tracking error ($\bm{s}^\text{T}\bm{s}$) is non-zero. Fig. \ref{fig.stateV} displays the tracking errors, presented as the partial Lyapunov function candidate $V_s=\frac{1}{2}\bm{s}^\text{T}\bm{s}$ (a reduced form of equation \eqref{eqn.lyapunov}), from all 30 flights. The outcomes are consistent with the observations from Figs. \ref{fig.hoveringflights110_trajectory_ef} and \ref{fig.hoveringflights110}. The tracking errors rapidly diminish in the beginning, before leveling off after nearly the first 30 s, owing to the contribution from the parameter adaptation. The obtained results reinforce the importance of the adaptive component to the tracking performance.

\subsection{Trajectory tracking flights} \label{sec.exp.traj}

To investigate the use of the proposed robot and strategy for transporting payloads, we performed trajectory tracking flights in the indoor setting with identical environments to the hovering flights.

\begin{figure}[t]
\psfrag{S1}[c][c][0.9]{~~S1}
\psfrag{S2}[c][c][0.9]{S2}
\psfrag{S3}[c][c][0.9]{S3}
\psfrag{S4}[c][c][0.9]{S4}
\psfrag{PA}[c][c][0.9]{Platform A}
\psfrag{PE}[c][c][0.9]{Platform E}
\psfrag{Xp}[c][c][0.9]{X position (cm)}
\psfrag{Yp}[c][c][0.9]{Y position (cm)}
\psfrag{Zp}[c][c][0.9]{Z position (cm)}
\psfrag{time}[c][c][0.9]{time (s)}
\centering
\includegraphics[width=90mm]{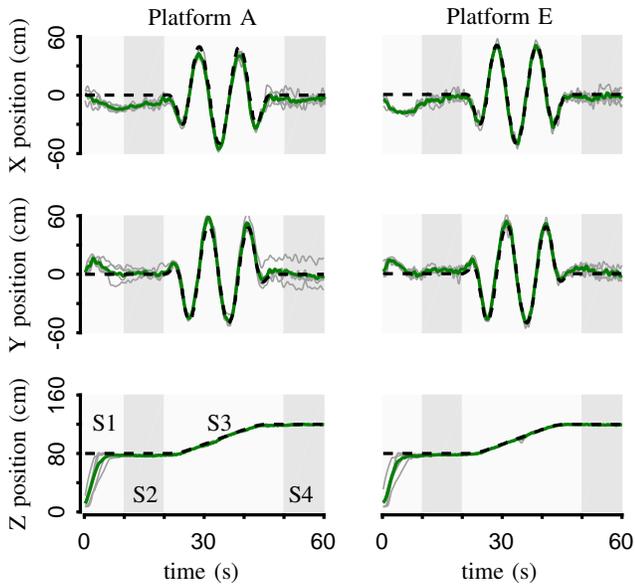}
\caption{Positions of Platforms A and E from stages S1 to S4 during five flights (gray lines) with respect to the desired trajectory (black dashed lines). The dark green lines are the average from all five flights.}\label{fig.traj_pos}
\end{figure}

The trajectory used for demonstration can be divided into four stages. This begins with a takeoff (S1: 0 s $\leq t \leq$ 10 s), followed by a 10-second hovering (S2: 10 s $\leq t \leq$ 20 s), a 30-second tracking of a helical trajectory with 50-cm radius (S3: 20 s $\leq t \leq$ 50 s), and another 10-second hovering before landing (S4: 50 s $\leq t \leq$ 60 s). The time-varying setpoints were generated based on $9^{\mbox{th}}$-order polynomial path primitives and sinusoidal functions. The yaw angle was set to be zero during the whole period. The setpoint trajectory is plotted as dashed lines in Fig. \ref{fig.traj_pos}.
\begin{figure}[ht]
\psfrag{PA}[c][c][0.9]{Platform A}
\psfrag{PE}[c][c][0.9]{Platform E}
\psfrag{Fstages}[c][c][0.9]{Flight stages}
\psfrag{Ax2}[c][c][0.9]{-9.8}
\psfrag{Ax3}[c][c][0.9]{-4.3}
\psfrag{Ax4}[c][c][0.9]{-4.7}
\psfrag{Ay2}[c][c][0.9]{0.9}
\psfrag{Ay3}[c][c][0.9]{2.8}
\psfrag{Ay4}[c][c][0.9]{-0.7}
\psfrag{Az2}[c][c][0.9]{-2.8}
\psfrag{Az3}[c][c][0.9]{-1.3}
\psfrag{Az4}[c][c][0.9]{-0.5}
\psfrag{Ex2}[c][c][0.9]{-4.7}
\psfrag{Ex3}[c][c][0.9]{-1.1}
\psfrag{Ex4}[c][c][0.9]{-2.6}
\psfrag{Ey2}[c][c][0.9]{3.1}
\psfrag{Ey3}[c][c][0.9]{2.8}
\psfrag{Ey4}[c][c][0.9]{1.8}
\psfrag{Ez2}[c][c][0.9]{-2.0}
\psfrag{Ez3}[c][c][0.9]{-0.6}
\psfrag{Ez4}[c][c][0.9]{-0.6}
\psfrag{Xe}[c][c][0.9]{X error (cm)}
\psfrag{Ye}[c][c][0.9]{Y error (cm)}
\psfrag{Ze}[c][c][0.9]{Z error (cm)}
\psfrag{FS2}[c][c][0.9]{S2}
\psfrag{FS3}[c][c][0.9]{S3}
\psfrag{FS4}[c][c][0.9]{S4}
\centering
\includegraphics[width=90mm]{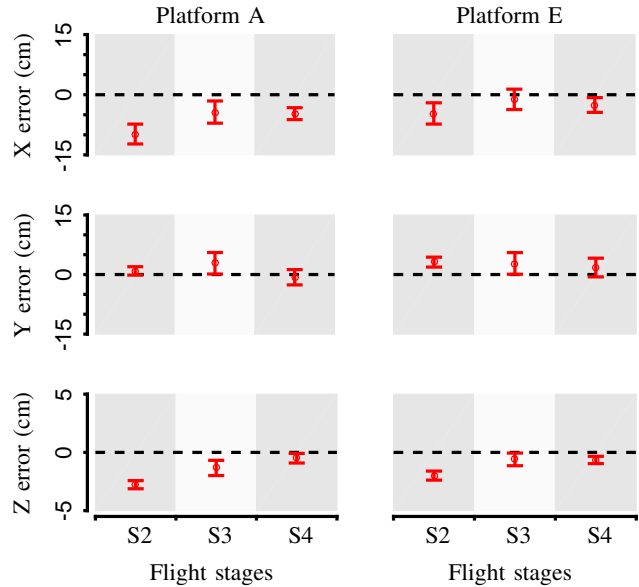}
\caption{The averages and standard deviations of the position errors of Platforms A and E from stages `S2' to `S4'. In flight stages `S2' and `S4', the robots were hovering while the robots were commanded to follow a spiral path in `S3'. The plots illustrate comparable flight performance of both platforms in different flight stages.}\label{fig.traj_avg_std}
\end{figure}

Platforms A and E were chosen for the experiments. Each platform carried out five flights with the adaptive part of the controller enabled from the beginning. The starting conditions were identical those used for sustained flight experiments.   In addition, we increased the adaptive gains from the hovering flight to compensate for a more complicated trajectory (refer to Table \ref{tab.parameters_tuning} in the Supplemental Materials). The initial 10-second hovering stage (S2) was intended for the robots to adapt its estimates before following the helical path. The experimental results are presented in Figs. \ref{fig.traj_pos} and \ref{fig.traj_avg_std}. Flight videos are available in the Supplemental Materials.

Fig. \ref{fig.traj_pos} shows the recorded trajectories of both platforms with respect to the setpoints. The averaged trajectories from all five flights are also plotted in dark green. The results suggest the robots were capable of accurately follow the prescribed trajectory. In more detail, Fig. \ref{fig.traj_avg_std} presents tracking errors of Platforms A and E in terms of the average errors and standard deviations. Three flight stages (S2-S4) are considered separately. It can be seen that the position errors are generally below 5 cm for both platforms. The errors of the tracking stage (S3) are comparable to that of the hovering stages (S2 and S4) and the previous hovering flights (Fig. \ref{fig.hoveringflights110}). Moreover, the errors show a decreasing trend over time. This is likely contributed by the adaptive component of the controller. Overall, the results verify that the proposed modular robots and associated control scheme is capable of transporting payloads between desired waypoints. 

\section{Conclusion and Discussion} \label{sec.conclusion}

We have proposed a novel modular design for a reconfigurable multirotor platform (UFOs). This system is composed of propelling modules and a control module that can be attached to a different payload, which serves as an airframe for the vehicle. 

The ability to reconfigure the vehicle by switching the payload and attachment points on-the-fly results in the change in dynamics of the robot, rendering it infeasible for the flight controller to stabilize the flight unless the model is comprehensively re-evalulated. To this end, we leverage multiple IMUs for estimating the configuration of the robot without requiring manual calculation. When combined with the developed adaptive controller, the parameters are comprehensively updated online according to the tracking errors.

To demonstrate the concept, we manufactured flight modules for six robots with different configurations and payloads. The initial estimates of system parameters of all prototypes are obtained via the proposed IMU-based estimation strategies. We have applied the adaptive geometric controller for hovering and trajectory tracking flights. It turns out that additional trimming was required to further refine the estimated robot's configuration in order to achieve a stable liftoff for platforms with highly irregular configurations. The subsequent experimental results show that, after parameter convergence, all prototypes approached the setpoint with average position errors of a few centimeters. The results confirm that the developed parameter estimations and adaptive controller are suitable for the proposed UFO platform.

The proposed strategy can be regarded as an alternative method for aerial transport. The modular design offers benefits on the reusability and adaptability. However, there still exist multiple limitations and unsolved challenges in the current work. These include the restrictions on the rigidity of the payload, the suitability of the surface material for repeatable attachment, and the geometry of the payload required to ensure that all thrusts are aligned. To alleviate the restriction on the surface material of the payload, it is essential to design and employ better hardware and attachment mechanism. To tackle the issue on the rigidity of the payload and the requirement of a flat surface for module attachments, one possibility is to exploit redundancy and equip the flight modules with an ability to tilt the propellers (either by a user or actively) as present in fully-actuated robots \cite{ryll20176d}. Such strategy must be accompanied by the development of novel flight control methods to deal with the case of unparalleled thrusts in the presence of uncertain parameters. These remain future research directions.

\section*{Acknowledgment}
This work was substantially supported by the Research Grants Council of the Hong Kong Special Administrative Region of China (grant number CityU-11274016).

The authors would like to express their gratitude towards Yi Hsuan Hsiao for the support on mechanical design and Jing Shu for the assistance in PCB board design.

\bibliographystyle{IEEEtran}
\bibliography{UFOs_TRO}

% Generated by IEEEtran.bst, version: 1.12 (2007/01/11)
\begin{thebibliography}{10}
\providecommand{\url}[1]{#1}
\csname url@samestyle\endcsname
\providecommand{\newblock}{\relax}
\providecommand{\bibinfo}[2]{#2}
\providecommand{\BIBentrySTDinterwordspacing}{\spaceskip=0pt\relax}
\providecommand{\BIBentryALTinterwordstretchfactor}{4}
\providecommand{\BIBentryALTinterwordspacing}{\spaceskip=\fontdimen2\font plus
\BIBentryALTinterwordstretchfactor\fontdimen3\font minus
  \fontdimen4\font\relax}
\providecommand{\BIBforeignlanguage}[2]{{%
\expandafter\ifx\csname l@#1\endcsname\relax
\typeout{** WARNING: IEEEtran.bst: No hyphenation pattern has been}%
\typeout{** loaded for the language `#1'. Using the pattern for}%
\typeout{** the default language instead.}%
\else
\language=\csname l@#1\endcsname
\fi
#2}}
\providecommand{\BIBdecl}{\relax}
\BIBdecl

\bibitem{mahony2012multirotor}
R.~Mahony, V.~Kumar, and P.~Corke, ``Multirotor aerial vehicles,'' \emph{IEEE
  Robotics and Automation magazine}, vol.~20, no.~32, 2012.

\bibitem{floreano2015science}
D.~Floreano and R.~J. Wood, ``Science, technology and the future of small
  autonomous drones,'' \emph{Nature}, vol. 521, no. 7553, pp. 460--466, 2015.

\bibitem{mur2017orb}
R.~Mur-Artal and J.~D. Tard{\'o}s, ``Orb-slam2: An open-source slam system for
  monocular, stereo, and rgb-d cameras,'' \emph{IEEE Transactions on Robotics},
  vol.~33, no.~5, pp. 1255--1262, 2017.

\bibitem{zhou2018agile}
D.~Zhou, Z.~Wang, and M.~Schwager, ``Agile coordination and assistive collision
  avoidance for quadrotor swarms using virtual structures,'' \emph{IEEE
  Transactions on Robotics}, vol.~34, no.~4, pp. 916--923, 2018.

\bibitem{vasarhelyi2018optimized}
G.~V{\'a}s{\'a}rhelyi, C.~Vir{\'a}gh, G.~Somorjai, T.~Nepusz, A.~E. Eiben, and
  T.~Vicsek, ``Optimized flocking of autonomous drones in confined
  environments,'' \emph{Science Robotics}, vol.~3, no.~20, p. eaat3536, 2018.

\bibitem{zhao2017deformable}
N.~Zhao, Y.~Luo, H.~Deng, and Y.~Shen, ``The deformable quad-rotor: design,
  kinematics and dynamics characterization, and flight performance
  validation,'' in \emph{2017 IEEE/RSJ International Conference on Intelligent
  Robots and Systems (IROS)}.\hskip 1em plus 0.5em minus 0.4em\relax IEEE,
  2017, pp. 2391--2396.

\bibitem{pounds2018safety}
P.~E. Pounds and W.~Deer, ``The safety rotor—an electromechanical rotor
  safety system for drones,'' \emph{IEEE Robotics and Automation Letters},
  vol.~3, no.~3, pp. 2561--2568, 2018.

\bibitem{ryll2015novel}
M.~Ryll, H.~H. B{\"u}lthoff, and P.~R. Giordano, ``A novel overactuated
  quadrotor unmanned aerial vehicle: Modeling, control, and experimental
  validation,'' \emph{IEEE Transactions on Control Systems Technology},
  vol.~23, no.~2, pp. 540--556, 2015.

\bibitem{mueller2016relaxed}
M.~W. Mueller and R.~D'Andrea, ``Relaxed hover solutions for multicopters:
  Application to algorithmic redundancy and novel vehicles,'' \emph{The
  International Journal of Robotics Research}, vol.~35, no.~8, pp. 873--889,
  2016.

\bibitem{antonelli2017adaptive}
G.~Antonelli, E.~Cataldi, F.~Arrichiello, P.~R. Giordano, S.~Chiaverini, and
  A.~Franchi, ``Adaptive trajectory tracking for quadrotor mavs in presence of
  parameter uncertainties and external disturbances,'' \emph{IEEE Transactions
  on Control Systems Technology}, vol.~26, no.~1, pp. 248--254, 2017.

\bibitem{mcgarey2013autokite}
P.~McGarey and S.~Saripalli, ``Autokite experimental use of a low cost
  autonomous kite plane for aerial photography and reconnaissance,'' in
  \emph{2013 International Conference on Unmanned Aircraft Systems
  (ICUAS)}.\hskip 1em plus 0.5em minus 0.4em\relax IEEE, 2013, pp. 208--213.

\bibitem{abaunza2017dual}
H.~Abaunza, P.~Castillo, A.~Victorino, and R.~Lozano, ``Dual quaternion
  modeling and control of a quad-rotor aerial manipulator,'' \emph{Journal of
  Intelligent \& Robotic Systems}, vol.~88, no. 2-4, pp. 267--283, 2017.

\bibitem{kessens2016versatile}
C.~C. Kessens, J.~Thomas, J.~P. Desai, and V.~Kumar, ``Versatile aerial
  grasping using self-sealing suction,'' in \emph{2016 IEEE International
  Conference on Robotics and Automation (ICRA)}.\hskip 1em plus 0.5em minus
  0.4em\relax IEEE, 2016, pp. 3249--3254.

\bibitem{foehn2017fast}
P.~Foehn, D.~Falanga, N.~Kuppuswamy, R.~Tedrake, and D.~Scaramuzza, ``Fast
  trajectory optimization for agile quadrotor maneuvers with a cable-suspended
  payload.'' in \emph{Robotics: Science and Systems}, 2017, pp. 1--10.

\bibitem{kim2018origami}
S.-J. Kim, D.-Y. Lee, G.-P. Jung, and K.-J. Cho, ``An origami-inspired,
  self-locking robotic arm that can be folded flat,'' \emph{Science Robotics},
  vol.~3, no.~16, p. eaar2915, 2018.

\bibitem{sreenath2013dynamics}
K.~Sreenath and V.~Kumar, ``Dynamics control and planning for cooperative
  manipulation of payloads suspended by cables from multiple quadrotor
  robots,'' \emph{rn}, vol.~1, no.~r2, p.~r3, 2013.

\bibitem{estrada2018forceful}
M.~A. Estrada, S.~Mintchev, D.~L. Christensen, M.~R. Cutkosky, and D.~Floreano,
  ``Forceful manipulation with micro air vehicles,'' \emph{Science Robotics},
  vol.~3, no.~23, p. eaau6903, 2018.

\bibitem{booth2018omniskins}
J.~W. Booth, D.~Shah, J.~C. Case, E.~L. White, M.~C. Yuen, O.~Cyr-Choiniere,
  and R.~Kramer-Bottiglio, ``Omniskins: Robotic skins that turn inanimate
  objects into multifunctional robots,'' \emph{Science Robotics}, vol.~3,
  no.~22, p. eaat1853, 2018.

\bibitem{seo2019modular}
J.~Seo, J.~Paik, and M.~Yim, ``Modular reconfigurable robotics,'' \emph{Annual
  Review of Control, Robotics, and Autonomous Systems}, 2019.

\bibitem{romanishin20153d}
J.~W. Romanishin, K.~Gilpin, S.~Claici, and D.~Rus, ``3{D} {M}-{B}locks:
  Self-reconfiguring robots capable of locomotion via pivoting in three
  dimensions,'' in \emph{2015 IEEE International Conference on Robotics and
  Automation (ICRA)}.\hskip 1em plus 0.5em minus 0.4em\relax IEEE, 2015, pp.
  1925--1932.

\bibitem{oung2014distributed}
R.~Oung and R.~D'Andrea, ``The distributed flight array: Design,
  implementation, and analysis of a modular vertical take-off and landing
  vehicle,'' \emph{The International Journal of Robotics Research}, vol.~33,
  no.~3, pp. 375--400, 2014.

\bibitem{saldana2018modquad}
D.~Saldana, B.~Gabrich, G.~Li, M.~Yim, and V.~Kumar, ``Modquad: the flying
  modular structure that self-assembles in midair,'' in \emph{2018 IEEE
  International Conference on Robotics and Automation (ICRA)}.\hskip 1em plus
  0.5em minus 0.4em\relax IEEE, 2018, pp. 691--698.

\bibitem{zhao2018design}
M.~Zhao, T.~Anzai, F.~Shi, X.~Chen, K.~Okada, and M.~Inaba, ``Design, modeling,
  and control of an aerial robot dragon: A dual-rotor-embedded multilink robot
  with the ability of multi-degree-of-freedom aerial transformation,''
  \emph{IEEE Robotics and Automation Letters}, vol.~3, no.~2, pp. 1176--1183,
  2018.

\bibitem{guerrier2012fault}
S.~Guerrier, A.~Waegli, J.~Skaloud, and M.-P. Victoria-Feser, ``Fault detection
  and isolation in multiple {MEMS}-{IMU}s configurations,'' \emph{IEEE
  Transactions on Aerospace and Electronic Systems}, vol.~48, no.~3, pp.
  2015--2031, 2012.

\bibitem{avram2015imu}
R.~C. Avram, X.~Zhang, J.~Campbell, and J.~Muse, ``{IMU} sensor fault diagnosis
  and estimation for quadrotor {UAV}s,'' \emph{IFAC--PapersOnLine}, vol.~48,
  no.~21, pp. 380--385, 2015.

\bibitem{rehder2016extending}
J.~Rehder, J.~Nikolic, T.~Schneider, T.~Hinzmann, and R.~Siegwart, ``Extending
  kalibr: Calibrating the extrinsics of multiple imus and of individual axes,''
  in \emph{2016 IEEE International Conference on Robotics and Automation
  (ICRA)}.\hskip 1em plus 0.5em minus 0.4em\relax IEEE, 2016, pp. 4304--4311.

\bibitem{mellinger2012trajectory}
D.~Mellinger, N.~Michael, and V.~Kumar, ``Trajectory generation and control for
  precise aggressive maneuvers with quadrotors,'' \emph{The International
  Journal of Robotics Research}, vol.~31, no.~5, pp. 664--674, 2012.

\bibitem{faessler2015automatic}
M.~Faessler, F.~Fontana, C.~Forster, and D.~Scaramuzza, ``Automatic
  re-initialization and failure recovery for aggressive flight with a monocular
  vision-based quadrotor,'' in \emph{2015 IEEE International Conference on
  Robotics and Automation (ICRA)}.\hskip 1em plus 0.5em minus 0.4em\relax IEEE,
  2015, pp. 1722--1729.

\bibitem{goodarzi2013geometric}
F.~Goodarzi, D.~Lee, and T.~Lee, ``Geometric nonlinear {PID} control of a
  quadrotor uav on {SE} (3),'' in \emph{2013 European Control Conference
  (ECC)}.\hskip 1em plus 0.5em minus 0.4em\relax IEEE, 2013, pp. 3845--3850.

\bibitem{faessler2017thrust}
M.~Faessler, D.~Falanga, and D.~Scaramuzza, ``Thrust mixing, saturation, and
  body-rate control for accurate aggressive quadrotor flight,'' \emph{IEEE
  Robotics and Automation Letters}, vol.~2, no.~2, pp. 476--482, 2017.

\bibitem{goodarzi2015geometric}
F.~A. Goodarzi, D.~Lee, and T.~Lee, ``Geometric adaptive tracking control of a
  quadrotor unmanned aerial vehicle on {SE} (3) for agile maneuvers,''
  \emph{Journal of Dynamic Systems, Measurement, and Control}, vol. 137, no.~9,
  p. 091007, 2015.

\bibitem{chirarattananon2016perching}
P.~Chirarattananon, K.~Y. Ma, and R.~J. Wood, ``Perching with a robotic insect
  using adaptive tracking control and iterative learning control,'' \emph{The
  International Journal of Robotics Research}, vol.~35, no.~10, pp. 1185--1206,
  2016.

\bibitem{mu2017adaptive}
B.~Mu, E.~H. Ng, and P.~Chirarattananon, ``Adaptive control for multirotor
  systems with completely uncertain dynamics,'' in \emph{2017 IEEE
  International Conference on Robotics and Biomimetics (ROBIO)}.\hskip 1em plus
  0.5em minus 0.4em\relax IEEE, 2017, pp. 2225--2230.

\bibitem{mellinger2011minimum}
D.~Mellinger and V.~Kumar, ``Minimum snap trajectory generation and control for
  quadrotors,'' in \emph{2011 IEEE International Conference on Robotics and
  Automation (ICRA)}.\hskip 1em plus 0.5em minus 0.4em\relax IEEE, 2011, pp.
  2520--2525.

\bibitem{morrell2018differential}
B.~Morrell, M.~Rigter, G.~Merewether, R.~Reid, R.~Thakker, T.~Tzanetos,
  V.~Rajur, and G.~Chamitoff, ``Differential flatness transformations for
  aggressive quadrotor flight,'' in \emph{2018 IEEE International Conference on
  Robotics and Automation (ICRA)}.\hskip 1em plus 0.5em minus 0.4em\relax IEEE,
  2018, pp. 1--7.

\bibitem{hsiao2018ceiling}
Y.~H. Hsiao and P.~Chirarattananon, ``Ceiling effects for surface locomotion of
  small rotorcraft,'' in \emph{2018 IEEE/RSJ International Conference on
  Intelligent Robots and Systems (IROS)}.\hskip 1em plus 0.5em minus
  0.4em\relax IEEE, 2018, pp. 6214--621oun9.

\bibitem{PX42016}
\emph{Pixhawk Pilot Support Package (PSP) User Guide, Version 2.0}, Pilot
  Engineering Group, The MathWorks, Inc., 2016.

\bibitem{ryll20176d}
M.~Ryll, G.~Muscio, F.~Pierri, E.~Cataldi, G.~Antonelli, F.~Caccavale, and
  A.~Franchi, ``6d physical interaction with a fully actuated aerial robot,''
  in \emph{2017 IEEE International Conference on Robotics and Automation
  (ICRA)}.\hskip 1em plus 0.5em minus 0.4em\relax IEEE, 2017, pp. 5190--5195.

\end{thebibliography}
\vspace{-8mm}
\begin{IEEEbiography}[{\includegraphics[width=1in,height=1.25in,clip,keepaspectratio]{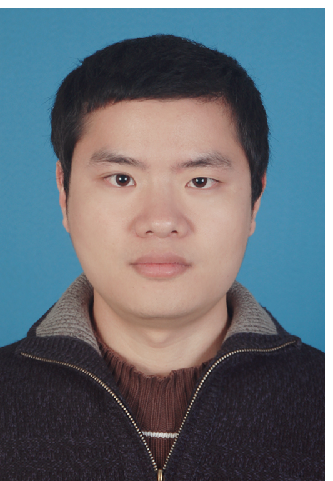}}]{Bingguo Mu}
received the B.A. degree in communication engineering from Nanchang University, China, in 2009, and the M.S. degree in signal and information processing from Sun Yat-sen University, in 2012. 

From 2012 to 2015, he was a software engineer in Samsung Electronics Guangzhou R$\&$D Center, Guangzhou, China. He is currently pursuing his Ph.D. degree with the Department of Biomedical Engineering, City University of Hong Kong, Kowloon, Hong Kong SAR, China. His current research interests include micro air vehicles, control theory and applications.
\end{IEEEbiography}
\vspace{-8mm}

\begin{IEEEbiography}[{\includegraphics[width=1in,height=1.25in,clip,keepaspectratio]{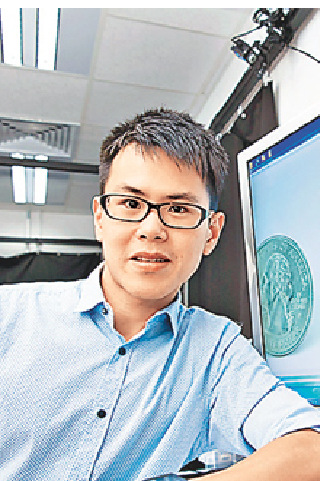}}]{Pakpong Chirarattananon}
(S'12-M'15) received the B.A. degree in natural sciences from the University of Cambridge, U.K., in 2009, and the Ph.D. degree in engineering sciences from Harvard University, Cambridge, MA, USA, in 2014.

He is currently an Assistant Professor with the Department of Biomedical Engineering, City University of Hong Kong, Kowloon, Hong Kong SAR, China. His research interests include bio-inspired robots, micro air vehicles, and the applications of control and dynamics in robotic systems.
\end{IEEEbiography}
\vfill

\clearpage
\onecolumn
\begin{center}
\textbf{\large Supplemental Materials for: }
\end{center}
\begin{center}
\textbf{Universal Flying Objects (UFOs): Modular Multirotor System for Flight of Rigid Objects}\\
\textbf{Bingguo Mu and Pakpong Chirarattananon}\\
\end{center}

\setcounter{equation}{0}
\setcounter{figure}{0}
\setcounter{table}{0}
\setcounter{page}{1}
\setcounter{section}{0}
\makeatletter
\renewcommand{\theequation}{S\arabic{equation}}
\renewcommand{\thefigure}{S\arabic{figure}}
\renewcommand{\thetable}{S\arabic{table}}
\renewcommand\thesection{S\Roman{section}}

\section{Trimming}\label{sec.trimming}

When the initial estimate of $\bm{A}$ is inaccurate, this leads to an excessively imbalanced torque that prevents the robot from lifting off stably. The purpose of the trimming process is to crudely adjust the $\hat{\bm{A}}$ based on the user's input from the remote control to lessen the imbalanced torques. With the following implementation, the trimming process is similar to those present in commercial products and can be carried out by a non-technical user. According to the experiments performed, trimming was only required for platforms with highly irregular configurations.

Suppose during the takeoff, the flight controller commands the nominal input $u_0$ calculated from $[g, \bm{0}]^\text{T}=\hat{\bm{A}}\bm{u}_{0}$ (refer to equation \eqref{eqn.dynamicmodel2}), in an attempt to produce zero net torque. Since, $\hat{\bm{A}}\neq\bm{A}$, this input $u_0$ results in some deviation ($\Delta g$, $\Delta \dot{\bm{\omega}}$) from the desired output according to
\begin{equation}
\left[
  \begin{array}{c}
    g+ \Delta g\\
    \Delta \dot{\bm{\omega}} \\
  \end{array}
\right]
=\bm{A}\bm{u}_0, \label{eqn.trim0}
\end{equation}
If the difference between $\bm{A}$ and $\hat{\bm{A}}$ is substantial, $ \Delta \dot{\bm{\omega}}$ is excessively large, the robot is unable to take off and crashes. The trimming process provides a crude update of $\hat{\bm{A}}$ in the form of $\Delta\hat{\bm{A}}$, for a quick update $\hat{\bm{A}}\rightarrow\hat{\bm{A}}+\Delta\hat{\bm{A}}$ that reduces the magnitude of the observed $ \Delta \dot{\bm{\omega}}$. One theoretically feasible solution can be obtained by solving
\begin{equation}
-\Delta \dot{\bm{\omega}}_i = \Delta \bm{\hat{A}}_{i+1} \bm{u}_0, \label{eqn.trim1}
\end{equation}
where the subscript $i\in\left\{ 1,2,3\right\} $ indicates the $i^{\text{th}}$ row of a vector/matrix. However, this is not feasible as $\Delta \dot{\bm{\omega}}$ is not directly measurable and the normal force from the ground is not considered. In practice, only the direction of $ \Delta \dot{\bm{\omega}}$ can be considered a meaningful measurement in this context.

To find a suitable $\Delta\hat{\bm{A}}$, we use the fact that all elements in $\bm{u}_0$ are positive. We choose
\begin{equation}
\Delta\hat{\bm{A}}_{i+1} = -\text{sgn}(\Delta \dot{\bm{\omega}}_i) \delta_i \text{abs}(\bm{\hat{A}}_{i+1}), \label{eqn.trim2}
\end{equation}
where $\delta_i$ is some positive step size and $\text{abs}(\cdot)$ is a element-wise absolute function. It can be seen that, there exists some $\delta_i$ that renders equation \eqref{eqn.trim2} a solution of \eqref{eqn.trim1}. In practice, the quantity $\text{sgn}(\Delta \dot{\bm{\omega}}_i) \delta_i$ is given by the operator's input via the RC controller. As long as the sign is correct, and $\delta_i$ is sufficiently small, the magnitude of the imbalanced torque is reduced.

The trimming method here only mitigates the imbalanced torque arisen from the poor estimate of $\hat{\bm{A}}$ only in this specific situation, the adjustment does not mean that make $\hat{\bm{A}}+\Delta\hat{\bm{A}}\rightarrow\bm{A}$. Unlike the adaptive controller, it also does not apply to more general flying conditions.

\newpage
\section{Supplemental Tables and Figures}
\begin{table}[ht]
\begin{center}
\caption{The mass of all prototypes.}
\begin{tabular}{ccc}\hline\hline%
& \multicolumn{2}{c}{\textbf{Mass (g)}} \\ \cline{2-3}%
\textbf{Prototypes} & \textbf{Actual} & \textbf{Estimate} \\ \hline%
\textbf{A} & $977.5$ & $970.0$ \\ \hdashline[0.1pt/2pt]%
\textbf{B} & $1265.3$ & $1330.0$ \\ \hdashline[0.1pt/2pt]%
\textbf{C} & $1651.3$ & $1690.0$ \\ \hdashline[0.1pt/2pt]%
\textbf{D} & $998.0$& $970.0$ \\ \hdashline[0.1pt/2pt]%
\textbf{E} & $1346.8$ & $1330.0$ \\ \hdashline[0.1pt/2pt]%
\textbf{F} & $2081.2$ & $1690.0$ \\ \hline%
\hline%
\end{tabular}\label{tab.Mass}
\end{center}
\end{table}

\begin{table*}[ht]
\begin{center}
\caption{The inertia tensor $\bm{I}$ (kg$\cdot$m$^2$) of all prototypes.}
\small{
\begin{tabular}{crrrrrr}\hline\hline%
\textbf{Platforms} & \multicolumn{3}{c}{\textbf{Values of $\bm{I}$ based on CAD estimate}} & \multicolumn{3}{c}{\textbf{Averaged estimate of $\bm{I}$}} \\ \hline %
\multirow{3}{*}{\textbf{A}}
& $7.29\times 10^{-3}$ & $-1.02\times 10^{-5}$ & $-2.52\times 10^{-5}$ & $7.56\times 10^{-3}$ & $-4.91\times 10^{-5}$ & $-1.71\times 10^{-4}$\\ %
& $-1.02\times 10^{-5}$ & $7.26\times 10^{-3}$ & $-1.18\times 10^{-5}$ & $-4.91\times 10^{-5}$ & $7.80\times 10^{-3}$ & $1.39\times 10^{-4}$\\ %
& $-2.52\times 10^{-5}$ & $-1.18\times 10^{-5}$ & $1.36\times 10^{-2}$ & $-1.71\times 10^{-4}$ & $1.39\times 10^{-4}$ & $1.32\times 10^{-2}$\\ \hdashline[0.1pt/2pt]% %
\multirow{3}{*}{\textbf{B}}
& $1.06\times 10^{-2}$ & $2.61\times 10^{-5}$ & $-4.77\times 10^{-5}$ & $1.07\times 10^{-2}$ & $4.98\times 10^{-4}$ & $-3.60\times 10^{-5}$\\ %
& $2.61\times 10^{-5}$ & $1.04\times 10^{-2}$ & $-2.30\times 10^{-5}$ & $4.98\times 10^{-4}$ & $1.04\times 10^{-2}$ & $-5.24\times 10^{-4}$\\ %
& $-4.77\times 10^{-5}$ & $-2.30\times 10^{-5}$ & $1.99\times 10^{-2}$ & $-3.60\times 10^{-5}$ & $-5.24\times 10^{-4}$ & $1.95\times 10^{-2}$\\ \hdashline[0.1pt/2pt]% %
\multirow{3}{*}{\textbf{C}}
& $2.09\times 10^{-2}$ & $1.67\times 10^{-4}$ & $1.77\times 10^{-5}$ & $1.90\times 10^{-2}$ & $-1.09\times 10^{-3}$ & $-1.63\times 10^{-3}$\\ %
& $1.67\times 10^{-4}$ & $2.12\times 10^{-2}$ & $-1.82\times 10^{-5}$ & $-1.09\times 10^{-3}$ & $1.80\times 10^{-2}$ & $-6.92\times 10^{-4}$\\ %
& $1.77\times 10^{-5}$ & $-1.82\times 10^{-5}$ & $4.09\times 10^{-2}$ & $-1.63\times 10^{-3}$ & $-6.92\times 10^{-4}$ & $3.46\times 10^{-2}$\\ \hdashline[0.1pt/2pt]% %
\multirow{3}{*}{\textbf{D}}
& $7.85\times 10^{-3}$ & $-1.15\times 10^{-3}$ & $-8.73\times 10^{-5}$ & $8.08\times 10^{-3}$ & $-1.66\times 10^{-3}$ & $-4.77\times 10^{-4}$\\ %
& $-1.15\times 10^{-3}$ & $6.39\times 10^{-3}$ & $-1.04\times 10^{-4}$ & $-1.66\times 10^{-3}$ & $6.96\times 10^{-3}$ & $-6.74\times 10^{-5}$\\ %
& $-8.73\times 10^{-5}$ & $-1.04\times 10^{-4}$ & $1.33\times 10^{-2}$ & $-4.77\times 10^{-4}$ & $-6.74\times 10^{-5}$ & $1.34\times 10^{-2}$\\ \hdashline[0.1pt/2pt]% %
\multirow{3}{*}{\textbf{E}}
& $9.98\times 10^{-3}$ & $3.46\times 10^{-4}$ & $-2.76\times 10^{-5}$ & $1.10\times 10^{-2}$ & $6.39\times 10^{-4}$ & $2.17\times 10^{-4}$\\ %
& $3.46\times 10^{-4}$ & $9.59\times 10^{-3}$ & $-1.07\times 10^{-4}$ & $6.39\times 10^{-4}$ & $1.06\times 10^{-2}$ & $2.48\times 10^{-4}$\\ %
& $-2.76\times 10^{-5}$ & $-1.07\times 10^{-4}$ & $1.37\times 10^{-2}$ & $2.17\times 10^{-4}$ & $2.48\times 10^{-4}$ & $1.44\times 10^{-2}$\\ \hdashline[0.1pt/2pt]% %
\multirow{3}{*}{\textbf{F}}
& $1.98\times 10^{-2}$ & $-3.02\times 10^{-4}$ & $-2.06\times 10^{-3}$ & $2.29\times 10^{-2}$ & $-2.31\times 10^{-3}$ & $1.69\times 10^{-3}$\\ %
& $-3.02\times 10^{-4}$ & $5.89\times 10^{-2}$ & $1.42\times 10^{-3}$ & $-2.31\times 10^{-3}$ & $5.41\times 10^{-2}$ & $3.36\times 10^{-3}$\\ %
& $-2.06\times 10^{-3}$ & $1.42\times 10^{-3}$ & $7.38\times 10^{-2}$ & $1.69\times 10^{-3}$ & $3.36\times 10^{-3}$ & $6.80\times 10^{-2}$\\
\hline\hline%
\end{tabular}}\label{tab.prototypes_inertia}
\end{center}
\end{table*}

\begin{table*}[ht]
\begin{center}
\caption{The configuration matrix $\bm{A}$ of platforms A-F.}
\small{
\begin{tabular}{ccrrrrrrrr}\hline\hline%
\textbf{Platforms} & \textbf{Estimates} & \multicolumn{8}{c}{\textbf{Configuration matrix $\bm{A}$ (in SI Units)}} \\\hline%
\multirow{8}{*}{\textbf{A}} & \multirow{4}{*}{\textbf{CAD estimate}}
& $1.02$ & $1.02$ & $1.02$ & $1.02$ & & & & \\ %
& & $0.25$ & $-26.21$ & $-0.11$ & $26.29$ & & & & \\ %
& & $-26.37$ & $0.88$ & $26.39$ & $0.27$ & & & & \\ %
& & $1.16$ & $-1.23$ & $1.21$ & $-1.13$ & & & & \\ \cdashline{2-10}[0.1pt/2pt]% %
&\multirow{4}{*}{\textbf{Averaged IMU-based estimate}}
& $1.03$ & $1.03$ & $1.03$ & $1.03$ & & & & \\ %
& & $1.48$ & $-25.75$ & $0.13$ & $26.42$ & & & & \\ %
& & $-26.48$ & $-0.51$ & $25.76$ & $1.53$ & & & & \\ %
& & $1.60$ & $-1.90$ & $0.99$ & $-0.63$ & & & & \\ \hline %
\multirow{8}{*}{\textbf{B}} & \multirow{4}{*}{\textbf{CAD estimate}}
& $0.79$ & $0.79$ & $0.79$ & $0.79$ & $0.79$ & $0.79$ & & \\ %
& & $0.22$ & $15.79$ & $15.63$ & $-0.03$ & $-15.60$ & $-15.28$ & & \\ %
& & $18.40$ & $9.00$ & $-9.36$ & $-18.41$ & $-8.99$ & $9.26$ & & \\ %
& & $0.83$ & $-0.76$ & $0.84$ & $-0.83$ & $0.76$ & $-0.84$ & & \\ \cdashline{2-10}[0.1pt/2pt]% %
& \multirow{4}{*}{\textbf{Averaged IMU-based estimate}}
& $0.75$ & $0.75$ & $0.75$ & $0.75$ & $0.75$ & $0.75$ & & \\ %
& & $-1.44$ & $15.59$ & $17.02$ & $1.45$ & $-14.49$ & $-14.99$ & & \\ %
& & $18.44$ & $9.76$ & $-9.06$ & $-18.18$ & $-7.63$ & $7.93$ & & \\ %
& & $1.30$ & $-0.49$ & $0.72$ & $-1.30$ & $0.55$ & $-0.73$ & & \\ \hline %
\multirow{8}{*}{\textbf{C}} & \multirow{4}{*}{\textbf{CAD estimate}}
& $0.61$ & $0.61$ & $0.61$ & $0.61$ & $0.61$ & $0.61$ & $0.61$ & $0.61$ \\ %
& & $11.10$ & $7.98$ & $0.19$ & $-7.69$ & $-10.90$ & $-7.88$ & $-0.08$ & $7.67$ \\ %
& & $-0.12$ & $-7.65$ & $-10.83$ & $-7.60$ & $0.11$ & $7.81$ & $10.99$ & $7.64$  \\ %
& & $0.40$ & $-0.40$ & $0.40$ & $-0.40$ & $0.40$ & $-0.40$ & $0.40$ & $-0.40$ \\ \cdashline{2-10}[0.1pt/2pt]% %
& \multirow{4}{*}{\textbf{Averaged IMU-based estimate}}
& $0.59$ & $0.59$ & $0.59$ & $0.59$ & $0.59$ & $0.59$ & $0.59$ & $0.59$ \\ %
& & $11.52$ & $8.81$ & $-0.56$ & $-8.37$ & $-10.60$ & $-7.73$ & $1.08$ & $7.20$ \\ %
& & $1.58$ & $-6.93$ & $-12.38$ & $-7.93$ & $-0.51$ & $8.36$ & $10.91$ & $7.29$ \\ %
& & $1.29$ & $-0.04$ & $0.04$ & $-1.26$ & $-0.30$ & $-0.74$ & $0.85$ & $0.27$ \\ \hline %
\multirow{8}{*}{\textbf{D}} & \multirow{4}{*}{\textbf{CAD estimate}}
& $1.00$ & $1.00$ & $1.00$ & $1.00$ & & & & \\ %
& & $-21.82$ & $21.56$ & $17.34$ & $-6.59$ & & & & \\ %
& & $12.76$ & $20.97$ & $-9.84$ & $-29.41$ & & & & \\ %
& & $1.17$ & $-0.91$ & $1.25$ & $-1.49$ & & & & \\ \cdashline{2-10}[0.1pt/2pt]% %
& \multirow{4}{*}{\textbf{Averaged IMU-based estimate}}
& $1.03$ & $1.03$ & $1.03$ & $1.03$ & & & & \\ %
& & $-20.95$ & $19.70$ & $14.89$ & $-8.12$ & & & & \\ %
& & $14.64$ & $22.45$ & $-7.90$ & $-27.86$ & & & & \\ %
& & $0.80$ & $0.02$ & $1.48$ & $-2.09$ & & & & \\ \hline%
\multirow{8}{*}{\textbf{E}} & \multirow{4}{*}{\textbf{CAD estimate}}
& $0.74$ & $0.74$ & $0.74$ & $0.74$ & $0.74$ & $0.74$ \\ %
& & $0.21$ & $13.52$ & $14.68$ & $0.09$ & $-13.77$ & $-14.87$ \\ %
& & $16.67$ & $7.98$ & $-7.97$ & $-17.03$ & $-8.15$ & $8.23$ \\ %
& & $1.31$ & $-1.09$ & $1.14$ & $-1.31$ & $1.09$ & $-1.14$ \\ \cdashline{2-10}[0.1pt/2pt]% %
& \multirow{4}{*}{\textbf{Averaged IMU-based estimate}}
& $0.75$ & $0.75$ & $0.75$ & $0.75$ & $0.75$ & $0.75$ \\ %
& & $-0.80$ & $13.73$ & $14.73$ & $1.71$ & $-13.71$ & $-15.19$ \\ %
& & $16.74$ & $8.35$ & $-7.55$ & $-16.35$ & $-7.64$ & $7.26$ \\ %
& & $1.76$ & $-1.53$ & $0.69$ & $-1.78$ & $1.55$ & $-0.68$ \\ \hline%
\multirow{8}{*}{\textbf{F}} & \multirow{4}{*}{\textbf{CAD estimate}}
& $0.48$ & $0.48$ & $0.48$ & $0.48$ & $0.48$ & $0.48$ & $0.48$ & $0.48$ \\ %
& & $9.05$ & $8.96$ & $3.23$ & $-7.05$ & $-8.09$ & $-8.15$ & $-7.02$ & $3.42$ \\ %
& & $1.54$ & $-1.31$ & $-5.01$ & $-4.68$ & $-1.27$ & $1.60$ & $4.55$ & $4.97$  \\ %
& & $0.43$ & $0.08$ & $0.38$ & $-0.31$ & $-0.01$ & $-0.46$ & $-0.07$ & $-0.19$ \\ \cdashline{2-10}[0.1pt/2pt]% %
& \multirow{4}{*}{\textbf{Averaged IMU-based estimate}}
& $0.59$ & $0.59$ & $0.59$ & $0.59$ & $0.59$ & $0.59$ & $0.59$ & $0.59$ \\ %
& & $9.94$ & $9.54$ & $3.22$ & $-6.50$ & $-7.57$ & $-7.80$ & $-4.81$ & $4.14$ \\ %
& & $2.10$ & $-1.41$ & $-6.43$ & $-5.73$ & $-1.65$ & $2.00$ & $5.33$ & $5.95$ \\ %
& & $-0.13$ & $-0.45$ & $0.39$ & $0.18$ & $0.51$ & $-0.09$ & $0.16$ & $-0.58$ \\ %
\hline\hline%
\end{tabular}}\label{tab.prototypes_A0}
\end{center}
\end{table*}

\begin{table}[ht]
\begin{center}
\caption{Controller gains used for all prototypes}
\begin{tabular}{lcc}\hline\hline%
\multirow{2}{*}{\textbf{Parameters}}& \multicolumn{2}{c}{\textbf{Values}} \\
\cline{2-3}
 & \textbf{Hovering Flights (\ref{sec.exp.hovering})} & \textbf{Trajectory-tracking Flights (\ref{sec.exp.traj})}\\ \hline%
$k_\text{d}$ & $4.0$ & $4.0$ \\ \hdashline[0.1pt/2pt]%
$k_\text{p}$ & $8.0$ & $8.0$ \\ \hdashline[0.1pt/2pt]%
$k$ & $4.0$ & $4.0$ \\ \hdashline[0.1pt/2pt]%
$k_\text{z}$ & $2.0$ & $2.0$ \\ \hdashline[0.1pt/2pt]%
$k_\phi$, $k_\theta$, $k_\psi$ & $9.0$ & $10.0$ \\ \hdashline[0.1pt/2pt]%
$\lambda_\text{z}$ & $1.2\times 10^{-2}$ & $3.5\times 10^{-2}$ \\ \hdashline[0.1pt/2pt]%
$\lambda_\phi$ & $3.5\times 10^{-2}$ & $1.1\times 10^{-1}$ \\ \hdashline[0.1pt/2pt]%
$\lambda_\theta$ & $3.5\times 10^{-2}$ & $1.3\times 10^{-1}$ \\ \hdashline[0.1pt/2pt]%
$\lambda_\psi$ & $3.5\times 10^{-3}$ & $1.0\times 10^{-2}$ \\ \hline%
\hline%
\end{tabular}\label{tab.parameters_tuning}
\end{center}
\end{table}
\begin{figure*}[b]
\psfrag{Xp}[c][c][0.85]{X position (cm)}
\psfrag{Yp}[c][c][0.85]{Y position (cm)}
\psfrag{Zp}[c][c][0.85]{Z position (cm)}
\psfrag{time}[c][c][0.85]{time (s)}
\centering
\subfigure[]{\includegraphics[width=80mm]{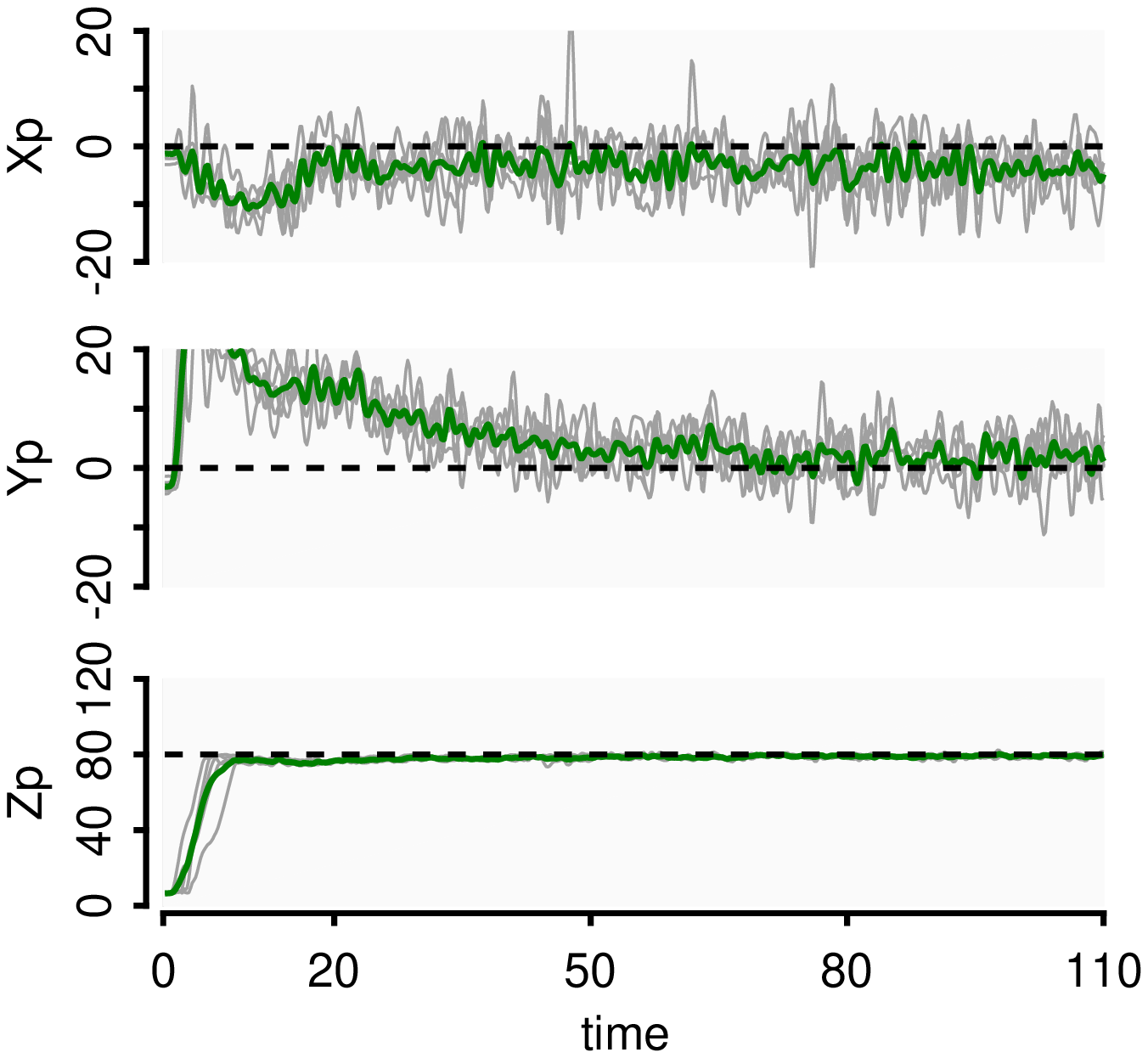}}
\subfigure[]{\includegraphics[width=80mm]{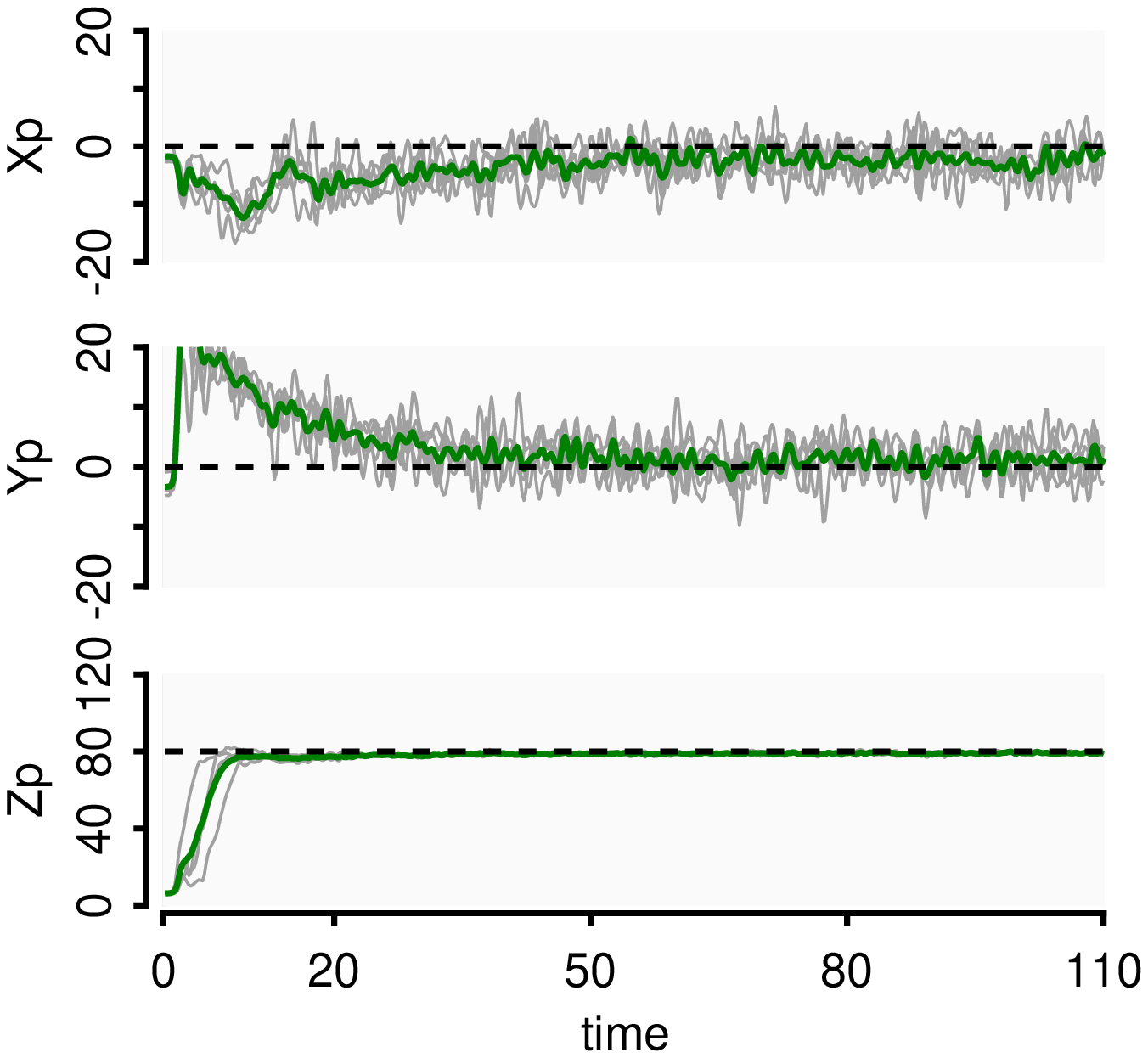}}
\subfigure[]{\includegraphics[width=80mm]{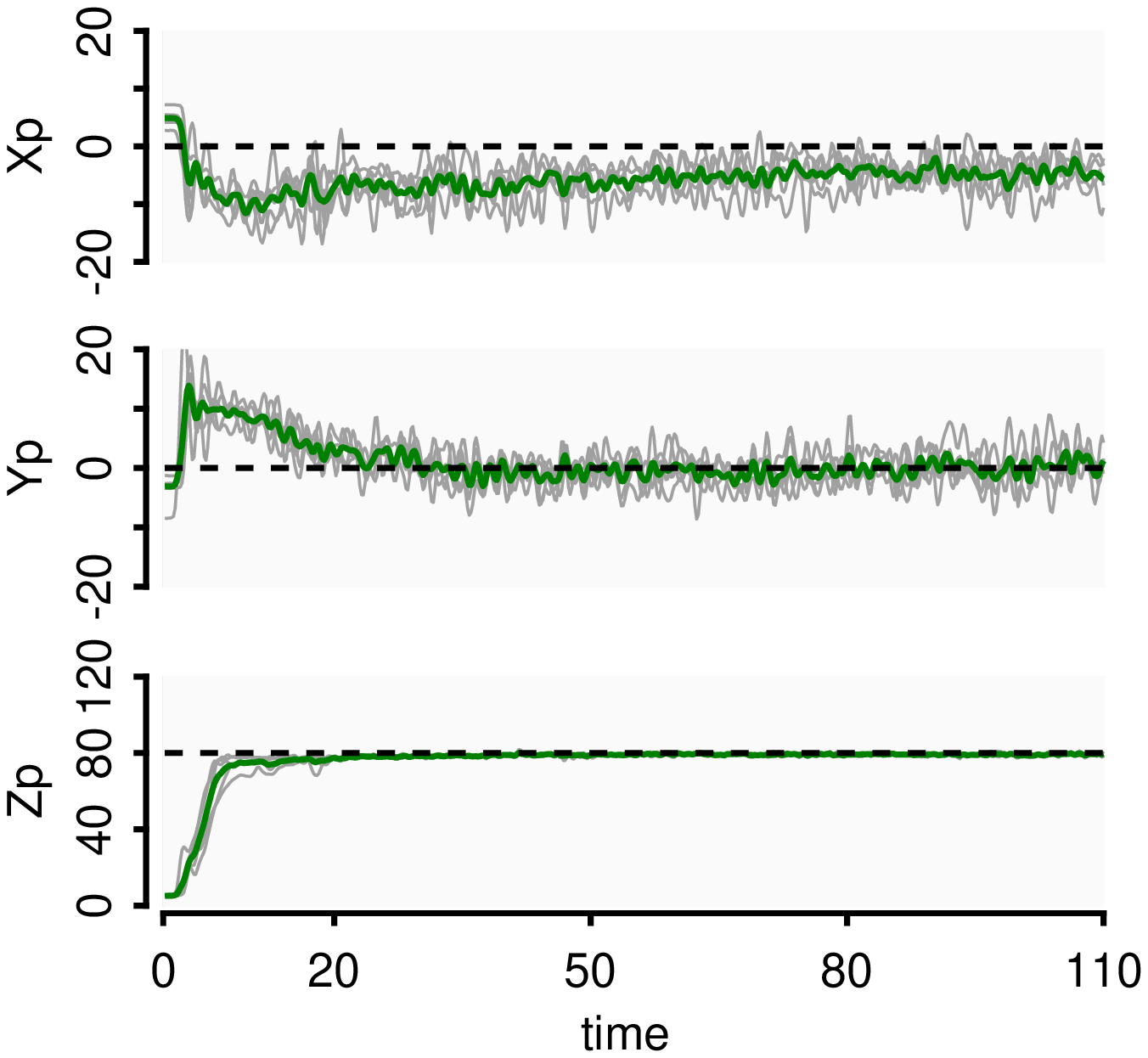}}
\subfigure[]{\includegraphics[width=80mm]{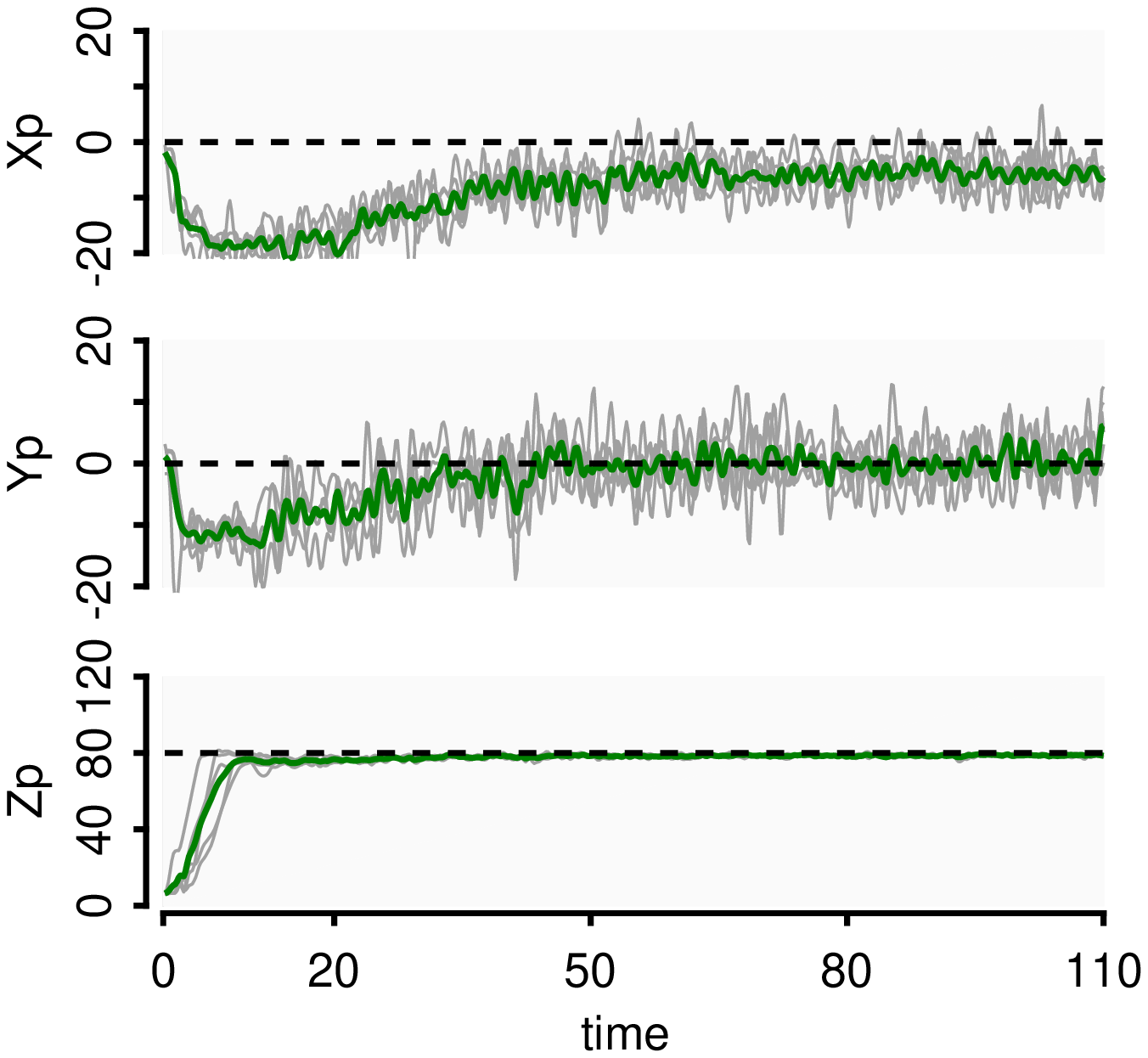}}
\caption{The trajectories of platforms (a) A, (b) B, (c) C and (d) D during five hovering flights (light green lines)  with respect to the setpoint (black dashed lines). The dark green lines are the averages from five flights.}\label{fig.hoveringflights110_trajectory_abcd}
\end{figure*}
\begin{table*}[ht]
\begin{center}
\caption{Table of Multimedia Extensions.}
\begin{tabular}{cccc}\hline\hline%
\textbf{Associated Files} & \textbf{Description} & \textbf{Type $\&$ Format} & \textbf{Size (MB)} \\ \hline%
Flight\_PlatformA.mp4 & Hovering flight of Platform A & Video/MP4 & 6.2 \\ \hdashline[0.1pt/2pt]%
Flight\_PlatformB.mp4 & Hovering flight of Platform B & Video/MP4 & 6.1 \\ \hdashline[0.1pt/2pt]%
Flight\_PlatformC.mp4 & Hovering flight of Platform C & Video/MP4 & 5.3 \\ \hdashline[0.1pt/2pt]%
Flight\_PlatformD.mp4 & Hovering flight of Platform D & Video/MP4 & 5.3 \\ \hdashline[0.1pt/2pt]%
Flight\_PlatformE.mp4 & Hovering flight of Platform E & Video/MP4 & 5.3 \\ \hdashline[0.1pt/2pt]%
Flight\_PlatformF.mp4 & Hovering flight of Platform F & Video/MP4 & 5.3 \\ \hdashline[0.1pt/2pt]%
Flight\_PlatformA\_traj.mp4 & Trajectory tracking flight of Platform A & Video/MP4 & 5.3 \\  \hdashline[0.1pt/2pt]%
Flight\_PlatformE\_traj.mp4 & Trajectory tracking flight of Platform E & Video/MP4 & 5.3 \\  \hdashline[0.1pt/2pt]%
UFO.mp4 & Complete demonstration & Video/MP4 & 4.5 \\ \hline%
\hline%
\end{tabular}\label{tab.vedios}
\end{center}
\end{table*}

\end{document}